\documentclass{article}

\usepackage{iclr2027_conference,times}

\usepackage[T1]{fontenc}
\usepackage[utf8]{inputenc}
\usepackage{microtype}

\usepackage{booktabs}
\usepackage{colortbl}
\usepackage{graphicx}
\usepackage{subcaption}
\usepackage{amssymb}
\usepackage{amsmath}
\usepackage{amsthm}
\usepackage{pifont}
\usepackage{enumitem}
\usepackage{multirow}
\usepackage{float}
\usepackage{placeins}
\usepackage{listings}
\usepackage[most]{tcolorbox}
\usepackage{tikz}
\usepackage{pgfplots}
\usepackage{wrapfig}
\usepackage{needspace}
\pgfplotsset{compat=1.18}
\usepgfplotslibrary{groupplots}
\usepackage{url}
\usepackage{hyperref}

\newcommand{\cmark}{\ding{51}}
\newcommand{\xmark}{\ding{55}}

\definecolor{codebg}{RGB}{248, 248, 248}
\definecolor{codeframe}{RGB}{200, 200, 200}
\definecolor{codekw}{RGB}{0, 112, 26}
\definecolor{codecomment}{RGB}{128, 128, 128}
\definecolor{codestr}{RGB}{163, 21, 21}

\lstdefinestyle{lean4}{
  language={},
  literate={×}{{$\times$}}1 {·}{{$\cdot$}}1,
  basicstyle=\ttfamily\small,
  backgroundcolor=\color{codebg},
  frame=single,
  rulecolor=\color{codeframe},
  framesep=6pt,
  xleftmargin=8pt,
  xrightmargin=8pt,
  breaklines=true,
  columns=fullflexible,
  keepspaces=true,
  morekeywords={theorem,def,let,by,do,return,where,import,open,namespace,end,if,then,else,match,with,fun,forall,exists},
  keywordstyle=\color{codekw}\bfseries,
  commentstyle=\color{codecomment}\itshape,
  morecomment=[l]{--},
  stringstyle=\color{codestr},
  showstringspaces=false,
  tabsize=2,
}
\lstdefinestyle{logexcerpt}{
  language={},
  basicstyle=\ttfamily\scriptsize,
  backgroundcolor=\color{gray!1},
  frame=none,
  breaklines=true,
  columns=fullflexible,
  keepspaces=true,
  showstringspaces=false,
  tabsize=2,
  aboveskip=1pt,
  belowskip=1pt,
}
\tcbset{
  casebox/.style={
    enhanced,
    colback=gray!2,
    colframe=gray!28,
    colbacktitle=gray!8,
    coltitle=black,
    fonttitle=\bfseries\small,
    fontupper=\small,
    boxrule=0.35pt,
    arc=1pt,
    left=4pt,
    right=4pt,
    top=3pt,
    bottom=3pt,
    before skip=3pt,
    after skip=3pt,
  }
}

\title{CktFormalizer: Autoformalization of Natural Language into Circuit Representations}

\author{
{\small Jing Xiong\thanks{Corresponding author: \texttt{junexiong@connect.hku.hk}},
Qi Han, Chenchen Ding, He Xiao, Zunhai Su, Xiachong Feng, Chaofan Tao, Ngai Wong} \\[4pt]
\multicolumn{1}{c}{The University of Hong Kong}
}

\iclrfinalcopy

\DeclareCaptionFormat{leftshiftedsub}{\hspace*{-6pt}\makebox[\linewidth][c]{#1#2#3}\par}
\begin{document}

\maketitle
\lhead{Preprint}

\begin{abstract}
Hardware infrastructure is a critical bottleneck for LLM-driven circuit design,
limiting what agents can express, compile, and iteratively refine within an
agentic loop. To address this bottleneck, we introduce \textsc{CktLean},
a typed hardware infrastructure embedded in Lean. It supports
hardware description, compilation to SystemVerilog, and interactive type-checking
and proof feedback through a persistent read-eval-print loop (REPL).
On this foundation, we build \textsc{CktFormalizer}, an agent framework for
hardware generation, repair, optimization, and source-level equivalence proving.
We evaluate structural correctness through
compilation, functional correctness through RTL and gate-level simulation,
and formal correctness through proofs relative to stated specifications.
Across VerilogEval, RTLLM, ResBench, and CVDP, \textsc{CktFormalizer} with
\textsc{CktLean} achieves compilation rates of 91.1\%--99.4\%. Among designs
that pass RTL simulation, 95.4\%--100.0\% jointly complete synthesis and
place-and-route and pass design-rule and layout-versus-schematic checks.
In a separate evaluation on 30 VerilogEval problems, interactive proof-state
feedback raises kernel-accepted equivalence proof completion from 53.3\% to 63.3\%.
Hardware evaluation feedback also guides architecture exploration and iterative
power, performance, and area (PPA) optimization, with synthesis-area reductions
of up to 58.6\% in the optimization loop. These results suggest that typed
representations and explicit proof-state feedback help structure the agentic loop:
compiler diagnostics guide targeted repairs, while proof-state feedback helps
agents identify what remains to be proved and determine the next proof step.
Project website: \url{https://ckt-formalizer.github.io/}.
\end{abstract}

\section{Introduction}
\label{sec:intro}

Large language models (LLMs) can translate natural-language hardware
specifications into register-transfer-level (RTL) designs written in
SystemVerilog~\citep{ieee1800_2023}, but syntactically valid RTL may still fail
to implement the intended
behavior~\citep{thakur2023benchmarking,liu2023verilogeval}.
Syntax checks establish conformance to the checked syntax rules, and RTL simulation tests finite traces;
neither guarantees equivalence to the intended specification or physical
realizability. Preserving the functional behavior is important
during iterative power, performance, and area (PPA) optimization, where repeated
design changes must remain correct through synthesis and physical implementation.

Higher-level HDLs such as Chisel~\citep{bachrach2012chisel},
Bluespec~\citep{nikhil2004bluespec}, and Clash~\citep{baaij2010clash}
provide abstractions that compile to RTL. However, addressing these correctness
challenges requires more than raising the abstraction level. We therefore use
a hardware DSL embedded in Lean~4~\citep{moura2021lean4} to combine hardware
description with machine-checked reasoning about circuit behavior.
\textsc{Sparkle}~\citep{hashimoto2026sparkle} provides a typed hardware representation
with SystemVerilog extraction, but leaves Mathlib's reusable lemmas and proof
tactics~\citep{mathlib4} underexplored for circuit equivalence proofs. We focus on
how an LLM agent can use these resources with interactive Lean feedback to
construct kernel-checked equivalence proofs. Hardware semantics and equivalence
statements must still be defined explicitly.
This motivates a hardware representation and interactive feedback interface
that support both circuit generation and equivalence proving.

\begin{tcolorbox}[width=\linewidth,colback=gray!10,boxrule=0pt,
  arc=2pt,boxsep=0pt,left=0pt,right=0pt,top=2pt,bottom=2pt,
  before skip=3pt,after skip=6pt,
  before upper={\parindent=0pt\parfillskip=0pt},after upper=\par]
\textbf{Research question.}
\emph{Can a typed hardware IR embedded in Lean, with interactive type-checking
and proof feedback, help agents generate correct hardware and prove
source-level equivalence?}
\end{tcolorbox}

\begin{wraptable}{r}{0.48\textwidth}
\centering
\small
\caption{SystemVerilog vs.\ \textsc{CktLean} within the supported generation subset.}
\vspace{-2mm}
\label{tab:compare-eda}
\setlength{\tabcolsep}{2pt}
\resizebox{\linewidth}{!}{%
\begin{tabular}{lcc}
\toprule
\textbf{Capability} & \textbf{SystemVerilog} & \textbf{CKTLEAN} \\
\midrule
Width consistency & \xmark & \cmark \\
Register-guarded feedback & \xmark & \cmark \\
Total case handling & \xmark & \cmark \\
Primary diagnostics & Tool logs & Type/macro errors \\
Kernel-checked proofs & \xmark & \cmark \\
Auto SV extraction & N/A & \cmark \\
\bottomrule
\end{tabular}
}
\end{wraptable}

To answer this question, we present \textsc{CktFormalizer}, an agent framework
for hardware generation, repair, optimization, and source-level equivalence proving.
Its foundation is our core contribution, \textbf{\textsc{CktLean}}, which
extends \textsc{Sparkle}'s typed IR, elaboration, and backend lowering.
Our analysis (\S\ref{sec:sparkle_limitations}) shows that \textsc{Sparkle}'s limited
support for parameterized state and memory, dynamic bit operations, signed
arithmetic, and multi-clock composition, together with compiler defects,
restricts the range of supported designs. \textsc{CktLean} addresses these gaps and
provides a persistent Lean read-eval-print loop (REPL).
\textsc{CktFormalizer} uses its compiler diagnostics and proof states in a
ReAct-style loop to repair hardware descriptions and develop equivalence proofs.
Table~\ref{tab:compare-eda} compares its representation and verification
capabilities with direct SystemVerilog generation.

We assess correctness along three dimensions: \emph{structural correctness},
assessed through type and pattern-match exhaustiveness checking; \emph{formal
correctness}, established at the source level relative to stated formal
specifications and assumptions through proofs accepted by Lean's kernel; and
\emph{functional correctness}, evaluated through RTL simulation, gate-level
simulation, and RTL equivalence checks (\S\ref{sec:correctness_taxonomy}).
Our contributions are as follows:
\begin{itemize}[leftmargin=*,itemsep=2pt,topsep=3pt]
    \item Our core contribution is \textbf{\textsc{CktLean}}, extending \textsc{Sparkle}
    to support more complex hardware designs and providing a persistent Lean
    REPL for interactive type-checking and proof feedback.
    \item We introduce \textsc{CktFormalizer}, which uses this infrastructure
    for type-guided hardware repair, synthesis-guided PPA optimization, and
    source-level equivalence proof generation.
    \item Across four benchmarks, all RTL-simulation-passing designs synthesize
    successfully, and 95.4--100\% complete place-and-route.
    Among RTL-simulation-passing designs, \textsc{CktFormalizer} with
    \textsc{CktLean} achieves higher gate-level simulation pass rates after
    synthesis and place-and-route than SystemVerilog generation using
    in-context learning or Verilog Agent on RTLLM and ResBench.
\end{itemize}

\section{Background and Related Work}
\label{sec:related}
\vspace{-2mm}
\paragraph{Typed Hardware Representation.}
Higher-level HDLs such as Chisel~\citep{bachrach2012chisel},
Bluespec~\citep{nikhil2004bluespec}, and C$\lambda$ash~\citep{baaij2010clash},
along with C/C++ HLS systems~\citep{cong2011hls,canis2011legup}, provide
abstractions that compile to RTL. \textsc{Sparkle}~\citep{hashimoto2026sparkle}
embeds hardware descriptions in Lean~4 using \texttt{Signal} combinators:
\texttt{Signal dom $\alpha$} represents a synchronous stream
$\mathbb{N} \to \alpha$ in clock domain \texttt{dom}.
Its \texttt{BitVec $N$} values encode width in the type; for example,
\texttt{Signal dom (BitVec 8)} produces an 8-bit value each cycle
(Figure~\ref{fig:lean_hdl_examples}). \textsc{Sparkle} uses Lean's type and exhaustiveness checks to reject ill-typed width
combinations and incomplete pattern matches. However, \textsc{Sparkle} has limited support for
parameterized state and memory, dynamic bit operations, signed arithmetic,
and multi-clock designs.
\begin{figure}[!htbp]
\centering
\begin{lstlisting}[style=lean4]
-- Example 1: 8-bit counter (sequential circuit)
def counter {dom : DomainConfig} : Signal dom (BitVec 8) :=
  Signal.circuit do
    let count <- Signal.reg 0#8
    count <~ count + 1#8
    return count

-- Example 2: Bit-width mismatch caught at compile time
def bad_add (a : Signal dom (BitVec 8))
            (b : Signal dom (BitVec 16)) : Signal dom (BitVec 8) :=
  a + b  -- ERROR: type mismatch (BitVec 8 vs BitVec 16)

\end{lstlisting}
\vspace{-4mm}
\caption{\textbf{Typed hardware representation in Lean.} An 8-bit counter uses explicit register feedback, while adding signals with incompatible bit widths is rejected by Lean's type checker.}
\label{fig:lean_hdl_examples}
\vspace{-3mm}
\end{figure}
\vspace{-3mm}
\paragraph{Lean Circuit Proofs and RTL Verification.}
\label{sec:compilation}
Generated RTL is validated through simulation, model checking, or
equivalence checking to assess whether it satisfies specified behavioral
properties or matches a reference
implementation~\citep{ieee1800_2023,clarke2018model,biere2009bounded,brayton2010abc}.
Proof assistants have also been applied to hardware verification: Kami and
Silver Oak use Coq to formally verify hardware designs, while ACL2 has been
used in industrial hardware verification~\citep{choi2017kami,silveroak2021,hunt2017industrial}.
\textsc{Sparkle} supports typed circuits and SystemVerilog extraction, but lacks
a general proof bridge between circuits and specifications and a proof of
extraction correctness~\citep{hashimoto2026sparkle}.
\textsc{CktLean} provides typed circuit representations that enable
\textsc{CktFormalizer} to formalize selected properties from natural-language
specifications and prove these properties and equivalence between circuit
implementations before and after optimization through iterative agent interaction
with the proof environment and proof-state feedback. Completed proofs are
checked by Lean's kernel
(Appendix~\ref{app:formal_verification}).

\vspace{-3mm}
\paragraph{RTL Generation and Agentic Workflows.}
Work on LLM-based hardware design ranges from
English-to-Verilog generation~\citep{pearce2020dave} and general or fine-tuned
code models~\citep{chen2021codex,roziere2023codellama,li2024starcoder2,thakur2024verigen}
to benchmarks such as VerilogEval~\citep{liu2023verilogeval} and
RTLLM~\citep{lu2024rtllm} and multi-step systems such as
Chip-Chat~\citep{blocklove2023chip} and ChipGPT~\citep{chang2024chipgpt}.
These approaches motivate examining which defects are exposed to the agent
during generation. SystemVerilog permits implicit width conversions,
incomplete combinational assignments, and feedback loops whose consequences
may evade finite RTL simulation and emerge during synthesis, timing checks,
or gate-level simulation.
\vspace{-3mm}
\paragraph{Agent Interactive Proving Workflows.}
Agent feedback loops allow a model to invoke compilers and testing tools,
observe their feedback, and iteratively revise its output. Tool-augmented
agents such as ReAct~\citep{yao2023react} and SWE-agent~\citep{yang2024swe}
support iterative editing and test feedback~\citep{jimenez2024swebench}, while
GPT-f~\citep{polu2020generative} and LeanDojo~\citep{yang2024leandojo}
demonstrate LLM-guided proof-step generation.
\textsc{CktFormalizer} uses \textsc{CktLean} and a persistent Lean REPL
for type-guided hardware repair and interactive equivalence proving.
The agent unfolds circuit definitions and iteratively proves behavioral
equivalence using proof-state feedback.

\FloatBarrier
\begin{figure}[!t]
    \centering
    \includegraphics[width=\linewidth]{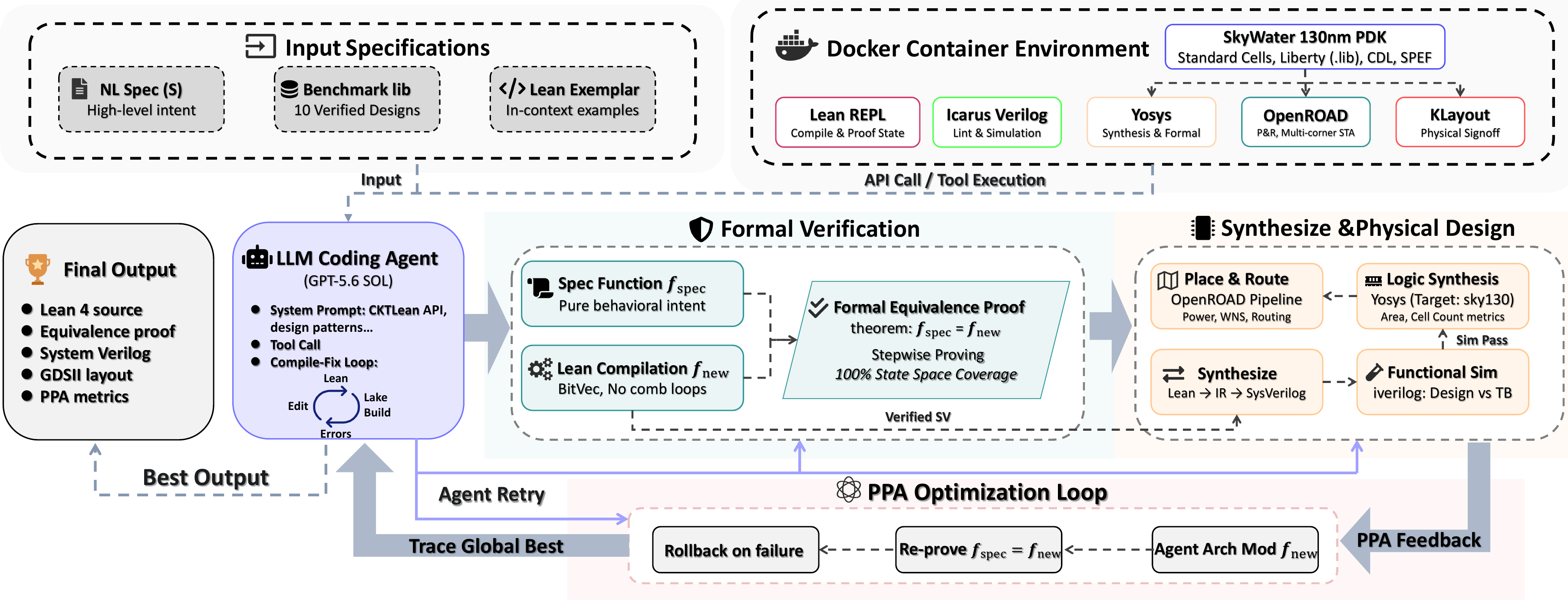}
    \caption{Overview of the \textsc{CktFormalizer} pipeline. Given a natural-language specification and examples, the agent uses compiler diagnostics and Lean REPL feedback to generate hardware and construct source-level equivalence proofs relative to stated specifications. Extracted SystemVerilog is evaluated through simulation, synthesis, and place-and-route using the SkyWater 130\,nm PDK. PPA feedback guides design revisions, with proof and evaluation feedback supporting acceptance or rollback.}
    \label{fig:pipeline}
\end{figure}
\vspace{-3mm}
\section{From \textsc{Sparkle} Limitations to CktLean}
\vspace{-2mm}
\label{sec:sparkle_limitations}
\label{sec:capability_gaps}
\begin{wraptable}{R}{0.43\textwidth}
\centering
\caption{Representative tasks by capability gap ($n$: task count; total 17).}
\label{tab:capability_gaps}
\vspace{-4pt}
\small
\setlength{\tabcolsep}{3pt}
\renewcommand{\arraystretch}{1.12}
\begin{tabular}{>{\raggedright\arraybackslash}p{\dimexpr0.43\linewidth-6pt\relax}>{\raggedright\arraybackslash}p{\dimexpr0.57\linewidth-22pt\relax}>{\centering\arraybackslash}p{10pt}}
\toprule
\textbf{Design class} & \textbf{Example tasks} & $n$ \\
\midrule
Stateful designs & Parking controller; restoring divider & 2 \\
\addlinespace[3pt]
Memories & Synchronous LIFO & 2 \\
\addlinespace[3pt]
Bit operations & Hamming encoder and decoder & 2 \\
\addlinespace[3pt]
Signed arithmetic & Complex multiplier; low-pass filter & 8 \\
\addlinespace[3pt]
Multiple clocks & Pulse synchronizer; asynchronous FIFO & 3 \\
\bottomrule
\end{tabular}
\end{wraptable}

We analyze the original \textsc{Sparkle} toolchain using 17 representative hardware tasks that expose
five gaps in design support and compilation (Table~\ref{tab:capability_gaps}).
Appendix~\ref{app:cktlean_extensions}, Table~\ref{tab:capability_task_details}, lists all task IDs and the corresponding capability limitations.
The tasks cover stateful designs, memories, bit operations, signed
arithmetic, and multiple clocks. These tasks help identify limitations in \textsc{Sparkle}'s hardware
representation and compilation pipeline.
\vspace{-3.5mm}
\paragraph{Stateful designs.}
Car parking and restoring division require derived widths, parameters,
loops, and tuple state to remain consistent during compilation.
The original elaboration path could lose symbolic width and parameter
dependencies when processing stateful circuits. This limits the ability to
represent configurable architectures faithfully: a circuit's state and
module interfaces must retain the width relationships specified by its
parameters throughout RTL generation.
\vspace{-4mm}
\paragraph{Memories.}
Sync LIFO illustrates the need to handle memory in stateful circuits in which
registers store the state used to compute the next state, including reset,
initialization, and read/write behavior. Memory width
and depth must remain available during lowering. Premature specialization
of these dimensions or incomplete handling of memory operations limits
parameterized memory support and requires separate checks of the emitted
RTL's behavior.
\vspace{-3mm}
\paragraph{Bit operations.}
Hamming-code encoder and decoder tasks combine indexed bit construction,
dynamic selection, and parity reduction. In the original compilation path,
symbolic dimensions and local aliases could be lost, incorrectly collapsing
vector expressions into scalar or fixed-width forms. Consequently, a
well-typed description need not retain its intended bit structure in the
emitted RTL.
\vspace{-3mm}
\paragraph{Signed arithmetic.}
QAM, comparators, complex multipliers, filters, Sobel, and adder tasks require
consistent sign extension, arithmetic shifts, truncation, and saturation.
Gaps in the representation and lowering of these operations restrict the
range of arithmetic designs that can be expressed and compiled faithfully.
In particular, correct widths alone do not establish that signed operations
have the intended numerical behavior across backends.
\vspace{-3mm}
\paragraph{Multiple clocks.}
Clock-domain-crossing pulse synchronizers and an asynchronous FIFO require distinct
clock domains, explicit clock/reset mappings, and cross-domain structure.
A single-clock representation cannot express these top-level relationships
adequately. This capability gap concerns circuit composition rather than
width parameterization and limits support for designs with multiple
synchronous domains.

\vspace{-1mm}
To address the five capability gaps identified above, \textsc{CktLean}
extends \textsc{Sparkle}'s typed IR and compilation pipeline to preserve parameter
dependencies, bit structure, signedness, and clock-domain information.
These extensions broaden support for complex hardware logic;
Appendix~\ref{app:cktlean_extensions} details their implementation.
Passing Lean's type checks and generating SystemVerilog do not guarantee
functional correctness: in the recorded parameterized-design run, the restoring
divider and synchronous LIFO passed these checks but failed simulation, with a
timeout in the former and incorrect full-flag behavior in the latter.
These failures motivate \textsc{CktFormalizer}'s agent workflow, which uses
compiler feedback to iteratively diagnose and repair generated designs.

\vspace{-2mm}
\section{Method}
\vspace{-3mm}
\label{sec:method}
\suppressfloats[t]
Figure~\ref{fig:pipeline} summarizes \textsc{CktFormalizer}'s workflow built on \textsc{CktLean}. To clarify what the workflow verifies at each stage, we first distinguish
three dimensions of correctness, then describe how the agent generates,
evaluates, and optimizes hardware.

\vspace{-3mm}
\subsection{Three Dimensions of Correctness}
\vspace{-1mm}
\label{sec:correctness_taxonomy}
We distinguish three dimensions of correctness: structural correctness of
hardware descriptions, formal correctness of circuit models, and functional
correctness of generated RTL relative to natural-language specifications.
\textsc{CktLean} extends the supported hardware representation and compilation
pipeline, while \textsc{CktFormalizer} uses compiler diagnostics, proof-state
feedback, and RTL verification to assess these dimensions during generation
and repair.

\emph{Structural correctness} means that a design is well formed within the
supported \textsc{CktLean} subset: bit widths, ports, and clock-domain parameters
are well typed, and pattern matches are exhaustive~\citep{hashimoto2026sparkle,moura2021lean4}.
Lean checks these properties during compilation (also supported by
\textsc{Sparkle} within its supported subset), and \textsc{CktFormalizer}
feeds compiler diagnostics back to the agent through the persistent Lean REPL
to guide repair.

\emph{Formal correctness} means constructing a lemma that states equivalence
between a \textsc{CktLean} circuit implementation and its formal specification,
or between two circuit implementations, and proving it in Lean. The lemma
expresses equality of their outputs for all inputs and execution steps covered
by its assumptions, and the proof is checked by Lean's
kernel~\citep{demoura2015lean,moura2021lean4}. Compiling a \textsc{CktLean} circuit
does not automatically establish its behavioral correctness.
For selected candidates, \textsc{CktFormalizer} guides the agent to develop these
proofs iteratively through the persistent Lean REPL provided by \textsc{CktLean},
including equivalence proofs between implementations before and after optimization.
Only completed proofs accepted without proof placeholders (\texttt{sorry} or \texttt{admit}) qualify. These proofs establish source-level correctness relative to formal specifications.
We separately assess the functional behavior of the generated RTL, as described below.

\emph{Functional correctness (evaluated by \textsc{CktFormalizer})} requires
the generated RTL to implement the intended behavior described by the
natural-language specification. \textsc{CktFormalizer} evaluates this behavior
through reference-testbench simulation and configured Yosys equivalence
checks~\citep{wolf2013yosys}. The evidence provided by these checks depends on
how faithfully the reference testbenches and RTL implementations capture the
original specification. Simulation covers only the exercised inputs and time
horizon, while RTL equivalence remains tool- and design-dependent.
Proofs of \textsc{CktLean} circuits do not automatically guarantee the behavior
of generated RTL, so we separately verify the emitted RTL within the agent loop.
\vspace{-3mm}
\subsection{Agentic Hardware Generation and Optimization Loop}
\vspace{-2mm}
\label{sec:agent_workflow}
The workflow combines compiler-guided hardware generation and repair with
evaluation, architecture exploration, and PPA optimization.
When enabled, interactive proof-state feedback supports source-level property
and equivalence proofs, while the generated RTL is evaluated separately.
\vspace{-3mm}
\paragraph{Coding Agent.}
\label{sec:agent}
The coding agent $\mathcal{A}$ is an LLM with tool use that iteratively writes,
type-checks, and repairs Lean HDL. We instantiate $\mathcal{A}$ with a
Codex-style coding harness (GPT-5.6-sol in the reported experiments).
The agent is initialized with a compact system prompt covering \textsc{CktLean}
idioms, common pitfalls, and a short generate-then-check workflow. The prompt
requires explicit registers to store the state used to compute the next state
in stateful circuits. The workspace includes generic Lean HDL example files covering
combinational logic, sequential circuits, and finite-state machines. Before generating a candidate, the agent may consult at most two
generic Lean HDL example files as in-context examples of coding patterns and API usage. The user
message contains the public natural-language specification and, when available,
the required port interface, but not a reference RTL body or the test harness.
\vspace{-3mm}
\paragraph{Iterative Compilation and Evaluation.}
\label{sec:eval_pipeline}
The agent generates a \textsc{CktLean} implementation from the natural-language
specification in an isolated workspace and iteratively repairs it using Lean
compiler diagnostics returned through the persistent Lean REPL. Generation and repair share a
budget of 100 turns per problem, where each turn comprises one LLM inference
and its tool calls. Initial generation is capped at 40 turns, with the remaining
budget available for compile-only repair. This repair
stage targets structural well-formedness and successful RTL extraction;
it does not use behavioral test failures to repair functional errors. The loop ends when compilation and
SystemVerilog extraction succeed or the budget is exhausted. An automated,
containerized evaluator then checks the emitted RTL through linting,
reference-testbench simulation, synthesis, and place-and-route to assess
functional behavior and obtain PPA metrics. Simulation assesses functional
correctness on the exercised traces, while synthesis and place-and-route
assess backend realizability. In the reported generation runs, testbench
results are not returned to the agent, and synthesis, place-and-route, and
proof-state-guided equivalence proving remain outside the generation-and-repair loop.
\vspace{-3mm}
\paragraph{Architecture Exploration and Metrics.}
\label{sec:arch_explore}
\label{sec:ppa_opt}
For designs with a large optimization space, the agent generates $C$
architectural candidates (default $C{=}3$), exploring choices such as parallel
versus serial datapaths and tree versus chain reductions. Eligible candidates
are compared using the evaluation and acceptance procedure below, and the
best candidate provides the starting point for iterative refinement.

We describe design quality using the metric vector
$\mathbf{m}_i=\bigl(A_i,P_i,W_i,N_i\bigr)$, whose components denote
area, power, WNS, and cell count, respectively.
Area is measured in $\mu\text{m}^2$ from Yosys synthesis statistics and includes
sequential and combinational contributions. Power is measured in
$\mu\text{W}$ using OpenROAD analysis across PVT corners. Worst negative slack
(WNS) is obtained from static timing analysis across SS, TT, and FF corners,
with $\text{WNS}=\min_p(T_{r,p}-T_{a,p})$,
where $p$ indexes timing paths, $T_{r,p}$ and $T_{a,p}$ denote required and
arrival times, and $\text{WNS}\geq 0$ indicates timing closure. Cell count serves as a proxy
for design complexity and routing congestion.
\vspace{-3mm}
\paragraph{Synthesis-Feedback-Guided PPA Optimization.}
Starting from the selected design, optimization runs for $K$ iterations
(default $K{=}3$), preserving the agent's conversation state. At iteration $i$,
the agent receives the available metrics, their history and changes from the
previous iteration, and guidance targeting the measured bottlenecks. It then
modifies the Lean source, for example by simplifying multiplexer logic,
removing redundant registers, or restructuring arithmetic expressions.

Candidate evaluation comprises recompilation, simulation, and synthesis;
place-and-route supplies power and timing metrics when included in the
configured evaluation flow. Recompilation checks that the modified design remains well typed
and can generate RTL, while simulation checks whether it still passes the testbench.
In the reported iterative optimization experiment (\S\ref{sec:ppa_opt_results}),
in-loop feedback uses Yosys area and cell count only; post-route metrics
are evaluated separately on the final design.
These synthesis metrics provide \emph{synthesis feedback} to guide subsequent Lean source revisions.

For candidates that pass evaluation, the full-metric acceptance rule compares
them with the best design found so far. Area, power, and cell count are minimized,
whereas WNS is maximized. For metric index $j\in\{1,2,3,4\}$, let
$d_j=+1$ for $j\in\{1,2,4\}$ and $d_3=-1$. We define the normalized change
$\Delta_j$ and accept a candidate if at least one metric improves by more than
$\tau$ and no metric degrades by more than $\epsilon$, where $\tau,\epsilon\geq 0$:
\begin{equation}
\Delta_j=d_j\frac{m'_j-m^*_j}{\max(|m^*_j|,s_j)},
\qquad
\bigl(\exists\,j:\Delta_j<-\tau\bigr)
\;\land\;
\bigl(\forall\,j:\Delta_j\leq\epsilon\bigr).
\label{eq:ppa_acceptance}
\end{equation}
Here $\mathbf{m}'$ and $\mathbf{m}^*$ denote the candidate and best-so-far metric
vectors, with $m'_j$ and $m^*_j$ their $j$th components. Each $s_j>0$ is a fixed, metric-specific normalization floor, expressed
in the same units as metric $j$ and set before optimization. These floors prevent
division by zero and stabilize comparisons when the best-so-far metric is close
to zero. With this convention, $\Delta_j<0$ always indicates improvement.
A candidate is retained only if it passes the configured evaluation checks and
acceptance criterion; otherwise, its changes are rolled back. After $K$
iterations, the best design found is restored as the final output.
\vspace{-2mm}
\paragraph{Formal Verification and Equivalence Checking.}
PPA optimization changes the circuit implementation, motivating equivalence
proofs between the original and revised \textsc{CktLean} designs. We express
both designs as Lean functions mapping input streams to output streams.
The equivalence lemma states that, under explicit assumptions on initialization
and reset, their outputs are equal for every input stream $x$ and clock cycle $t$:
\[
\forall x,t,\quad H(x,t)\rightarrow
\operatorname{orig}(x,t)=\operatorname{revised}(x,t),
\]
where $H$ collects the stated assumptions and $\operatorname{orig}(x,t)$ and
$\operatorname{revised}(x,t)$ denote the two designs' outputs at cycle $t$.
The persistent Lean REPL supports this proof-development process: in a ReAct-style loop, the
agent submits a proof step, inspects the resulting goals and diagnostics, and
selects or revises the next step. A completed, kernel-checked proof establishes
that the two source-level circuit models produce identical outputs under the
stated assumptions.

\FloatBarrier
\vspace{-3mm}
\section{Experiments}
\vspace{-2mm}
\label{sec:experiments}

To answer our research question, we evaluate:
\textit{(i)}~structural, formal, and functional correctness, comparing with direct Verilog baselines where applicable (\S\ref{sec:main_results});
\textit{(ii)}~architecture exploration and iterative PPA optimization using synthesis feedback (\S\ref{sec:optimization_verification}); and
\textit{(iii)}~source-level equivalence proof generation with interactive Lean feedback (\S\ref{sec:optimization_verification}).
\vspace{-2mm}
\subsection{Experimental Setup}
\label{sec:setup}
\vspace{-2mm}
\paragraph{Benchmarks.}
We evaluate on VerilogEval~\citep{liu2023verilogeval}, comprising 156 HDLBits
problems from wire assignments to complex FSMs; RTLLM~\citep{lu2024rtllm}, with
50 arithmetic, memory, and control designs; ResBench~\citep{guo2025resbench},
with 56 arithmetic, signal-processing, control, and datapath problems; and a
168-problem evaluation subset of Comprehensive Verilog Design Problems
(CVDP)~\citep{pinckney2025cvdp} covering FSM/control, arithmetic
datapaths, interconnects, and memory, which introduces more challenging tasks
reflecting real-world RTL design and verification scenarios.
We evaluate generated RTL in Docker containers using
OpenROAD~\citep{ajayi2019openroad} and the SkyWater 130\,nm
PDK~\citep{edwards2004sky130}, checking downstream realizability through
synthesis, place-and-route, design-rule checking (DRC), layout-versus-schematic
(LVS) verification, and gate-level simulation.
\vspace{-3mm}
\paragraph{Baselines.}
We compare \textsc{CktFormalizer} with four baselines:
RTLCoder~\citep{liu2024rtlcoder}, using the 7B RTLCoder-DeepSeek-v1.1 model;
CodeV~\citep{zhao2025codev}, using CodeV-All-QC;
ICL and Verilog Agent, both of which generate SystemVerilog directly.
ICL uses in-context learning (ICL) in a single model call. Verilog Agent is
our control baseline, which directly generates SystemVerilog and uses an
agent harness to iteratively repair the code with compiler feedback. It uses
the same model and 100-turn budget as \textsc{CktFormalizer}.
ICL, Verilog Agent, and \textsc{CktFormalizer} all use
\texttt{gpt-5.6-sol}. Whereas Verilog Agent repairs SystemVerilog directly,
\textsc{CktFormalizer} generates Lean HDL and uses Lean type-checking and
extraction diagnostics for iterative repair before emitting SystemVerilog
(\S\ref{sec:eval_pipeline}). For the \textsc{CktFormalizer} runs on VerilogEval
and RTLLM, reference implementation bodies are removed from the generation
prompts. All methods are evaluated using the same physical-design flow with
the SkyWater 130\,nm HD standard-cell library.

\subsection{Main Results}
\label{sec:main_results}
\vspace{-2mm}
Table~\ref{tab:main_results} summarizes the end-to-end results on VerilogEval, RTLLM, ResBench, and CVDP.
Downstream pass rates can exceed Sim Pass because they are conditioned on
passing RTL simulation, rather than evaluated over all problems.
\vspace{-4mm}
\begin{table}[t]
\centering
\captionsetup{skip=4pt}
\caption{End-to-end pass rates (\%): $100\times$ passing designs / evaluated designs. Denominators: all problems for Compile/Sim Pass; RTL-simulation-passing designs otherwise. Synth/P\&R: synthesis/place-and-route completion; DRC/LVS: design-rule/layout-versus-schematic check passes; GLS-Synth/GLS-P\&R: gate-level simulation passes after the respective stages. PD Pass: joint Synth, P\&R, DRC, and LVS success. Verilog Agent is our compiler-feedback control baseline; it and \textsc{CktFormalizer} each receive 100 turns per problem. These turns comprise initial generation, compile-only repair from compiler diagnostics, and tool use (edits and builds).}
\label{tab:main_results}
\small
\setlength{\tabcolsep}{3pt}
\resizebox{\columnwidth}{!}{%
\definecolor{pdpassbg}{RGB}{237,243,248}
\begin{tabular}{lllcccccccc>{\columncolor{pdpassbg}}c}
\toprule
\textbf{Benchmark} & \textbf{Method} & \textbf{HDL} & \textbf{Compile} & \textbf{Sim Pass} & \textbf{Synth} & \textbf{P\&R} & \textbf{DRC} & \textbf{LVS} & \textbf{GLS-Synth} & \textbf{GLS-P\&R} & \textbf{PD Pass} \\
\midrule
\multirow{7}{*}{\shortstack[l]{VerilogEval\\(156 problems)}} & RTLCoder & SystemVerilog & 82.1\% & 43.6\% & \textbf{100.0\%} & 97.1\% & 97.1\% & 97.1\% & \textbf{100.0\%} & \textbf{97.1\%} & 97.1\% \\
 & CodeV & SystemVerilog & 91.7\% & 42.3\% & 98.5\% & 95.5\% & 95.5\% & 95.5\% & 98.5\% & 95.5\% & 95.5\% \\
 & Verilog Agent & SystemVerilog & 98.1\% & \textbf{94.2\%} & \textbf{100.0\%} & 95.9\% & 95.9\% & 95.9\% & 96.6\% & 93.9\% & 95.9\% \\
\cmidrule(lr){2-12}
 & \multirow{2}{*}{ICL} & SystemVerilog & 98.1\% & \textbf{94.2}\% & \textbf{100.0\%} & 96.2\% & 96.2\% & 96.2\% & 97.0\% & 94.7\% & 96.2\% \\
 & & \textsc{CktLean} & 89.7\% & 40.4\% & 90.5\% & 85.7\% & 85.7\% & 85.7\% & 85.7\% & 85.7\% & 86.6\% \\
\cmidrule(lr){2-12}
 & \multirow{2}{*}{\textsc{CktFormalizer}} & \textsc{Sparkle} & 93.6\% & 85.9\% & \textbf{100.0\%} & \textbf{99.3\%} & \textbf{99.3\%} & \textbf{99.3\%} & 96.3\% & 97.0\% & \textbf{99.3\%} \\
 & & \textsc{CktLean} & \textbf{99.4\%} & 83.3\% & \textbf{100.0\%} & 95.4\% & 95.4\% & 95.4\% & 97.7\% & 94.6\% & 95.4\% \\
\midrule
\multirow{7}{*}{\shortstack[l]{RTLLM\\(50 problems)}} & RTLCoder & SystemVerilog & 70.0\% & 22.0\% & 90.9\% & 90.9\% & 90.9\% & 90.9\% & 81.8\% & 81.8\% & 90.9\% \\
 & CodeV & SystemVerilog & 80.0\% & 32.0\% & \textbf{100.0\%} & 93.8\% & 93.8\% & 93.8\% & 81.2\% & 81.2\% & 93.8\% \\
 & Verilog Agent & SystemVerilog & 94.0\% & \textbf{74.0\%} & 97.3\% & 97.3\% & 97.3\% & 97.3\% & 81.1\% & 78.4\% & 97.3\% \\
\cmidrule(lr){2-12}
 & \multirow{2}{*}{ICL} & SystemVerilog & \textbf{100.0}\% & \textbf{74.0\%} & 97.1\% & 97.1\% & 97.1\% & 97.1\% & 82.9\% & 80.0\% & 97.1\% \\
  & & \textsc{CktLean} & 96.0\% & 60.0\% & 93.3\% & 93.3\% & 93.3\% & 93.3\% & 93.3\% & 93.3\% & 93.3\% \\
\cmidrule(lr){2-12}
 & \multirow{2}{*}{\textsc{CktFormalizer}} & \textsc{Sparkle} & 82.0\% & 52.0\% & \textbf{100.0\%} & \textbf{100.0\%} & \textbf{100.0\%} & \textbf{100.0\%} & 84.6\% & \textbf{88.5\%} & \textbf{100.0\%} \\
 & & \textsc{CktLean} & \textbf{98.0\%} & 68.0\% & \textbf{100.0\%} & \textbf{100.0\%} & \textbf{100.0\%} & \textbf{100.0\%} & \textbf{88.2\%} & 88.2\% & \textbf{100.0\%} \\
\midrule
\multirow{7}{*}{\shortstack[l]{ResBench\\(56 problems)}} & RTLCoder & SystemVerilog & 85.7\% & 51.8\% & 93.1\% & 93.1\% & 89.7\% & 93.1\% & 93.1\% & 93.1\% & 89.7\% \\
 & CodeV & SystemVerilog & \textbf{100.0\%} & 50.0\% & 92.9\% & 92.9\% & 92.9\% & 92.9\% & 92.9\% & 92.9\% & 92.9\% \\
 & Verilog Agent & SystemVerilog & \textbf{100.0\%} & 83.9\% & 57.4\% & 57.4\% & 57.4\% & 57.4\% & 55.3\% & 55.3\% & 57.4\% \\
\cmidrule(lr){2-12}
 & \multirow{2}{*}{ICL} & SystemVerilog & \textbf{100.0\%} & 84.0\% & 60.5\% & 60.5\% & 60.5\% & 60.5\% & 58.1\% & 58.1\% & 60.5\% \\
 & & \textsc{CktLean} & 85.7\% & 73.2\% & 80.5\% & 80.5\% & 80.5\% & 80.5\% & 80.5\% & 80.5\% & 80.5\% \\
\cmidrule(lr){2-12}
 & \multirow{2}{*}{\textsc{CktFormalizer}} & \textsc{Sparkle} & \textbf{100.0\%} & 76.8\% & 95.3\% & 95.3\% & 95.3\% & 95.3\% & 93.0\% & 93.0\% & 95.3\% \\
 & & \textsc{CktLean} & 98.2\% & 78.6\% & \textbf{100.0\%} & \textbf{100.0\%} & \textbf{100.0\%} & \textbf{100.0\%} & \textbf{100.0\%} & \textbf{100.0\%} & \textbf{100.0\%} \\
\midrule
\multirow{7}{*}{\shortstack[l]{CVDP\\(168 problems)}} & RTLCoder & SystemVerilog & 85.7\% & 7.7\% & \textbf{100.0\%} & 92.3\% & 92.3\% & 92.3\% & 0.0\% & 0.0\% & 92.3\% \\
 & CodeV & SystemVerilog & 50.0\% & 4.2\% & 85.7\% & 85.7\% & 85.7\% & 85.7\% & 0.0\% & 0.0\% & 85.7\% \\
 & Verilog Agent & SystemVerilog & 98.8\% & \textbf{50.0\%} & 89.3\% & 88.1\% & 88.1\% & 88.1\% & 29.8\% & 29.8\% & 88.1\% \\
\cmidrule(lr){2-12}
 & \multirow{2}{*}{ICL} & SystemVerilog & 98.8\% & \textbf{50.0\%} & 89.7\% & 88.5\% & 88.5\% & 88.5\% & \textbf{33.3\%} & \textbf{33.3\%} & 88.5\% \\
 & & \textsc{CktLean} & 75.6\% & 16.1\% & 55.6\% & 55.6\% & 55.6\% & 55.6\% & 29.6\% & 25.9\% & 55.6\% \\
\cmidrule(lr){2-12}
 & \multirow{2}{*}{\textsc{CktFormalizer}} & \textsc{Sparkle} & 77.4\% & 25.6\% & 67.4\% & 67.4\% & 67.4\% & 67.4\% & 27.9\% & 27.9\% & 67.4\% \\
 & & \textsc{CktLean} & 91.1\% & 40.5\% & \textbf{100.0\%} & \textbf{97.1\%} & \textbf{97.1\%} & \textbf{97.1\%} & 30.9\% & 27.9\% & \textbf{97.1\%} \\
\bottomrule

\end{tabular}%
}
\end{table}

\paragraph{Structural Correctness.}
Lean compilation checks the supported structural invariants, including
bit-width consistency and exhaustive pattern matching
(\S\ref{sec:correctness_taxonomy}).
\textit{(i) Compilation.} \textsc{CktFormalizer} enables LLMs to generate
compilable hardware through \textsc{CktLean}'s typed, high-level representation,
achieving compilation rates of 99.4\%, 98.0\%, 98.2\%, and 91.1\% on
VerilogEval, RTLLM, ResBench, and CVDP, respectively. These results demonstrate
that high compilation coverage is achievable while enforcing the supported
structural constraints, making typed hardware generation a practical
alternative to direct SystemVerilog generation.
\textit{(ii) Backend realizability.} All RTL-simulation-passing designs from
\textsc{CktFormalizer} with \textsc{CktLean} synthesize successfully across
the four benchmarks. Among those same designs, P\&R, DRC, and LVS pass rates
are each 95.4\%, 100.0\%, 100.0\%, and 97.1\% on VerilogEval, RTLLM,
ResBench, and CVDP, respectively. The largest synthesis gap relative to the
SystemVerilog baselines occurs on ResBench, where ICL with SystemVerilog and
Verilog Agent reach 60.5\% and 57.1\%.
\textit{(iii) Infrastructure contribution.} Within the same
\textsc{CktFormalizer} workflow, replacing \textsc{Sparkle} with
\textsc{CktLean} raises synthesis pass rates from 95.3\% to 100.0\% on
ResBench and from 67.4\% to 100.0\% on CVDP. PD Pass also rises from 95.3\%
to 100.0\% and from 67.4\% to 97.1\%, respectively. These results support
\textsc{CktLean}'s contribution to backend realizability on these benchmarks.
\begin{figure}[!t]
\centering
\definecolor{resultblue}{HTML}{4C78A8}
\definecolor{resultorange}{HTML}{C67C3C}
\definecolor{resultteal}{HTML}{42958C}
\definecolor{resultgray}{HTML}{87939C}
\pgfplotsset{resultaxis/.style={
    width=0.62\linewidth,height=2.0cm,scale only axis,
    axis lines*=left,axis line style={black!35},
    tick style={black!35},tick align=outside,
    tick label style={font=\scriptsize},
    label style={font=\scriptsize},
    xlabel style={font=\scriptsize},
    ylabel style={font=\scriptsize},
    ymajorgrids=true,grid style={black!8},
    scaled ticks=false,
    legend style={font=\tiny,at={(0.5,1.06)},anchor=south,
        draw=none,fill=none,inner sep=0.5pt,legend columns=2,column sep=3pt,
        cells={anchor=west,text depth=0pt,text height=7pt}},
}}
\captionsetup[subfigure]{font=small,justification=centering,singlelinecheck=true,margin=0pt,skip=2pt}
\begin{subfigure}[t]{0.24\textwidth}
\centering
\begin{tikzpicture}
\begin{axis}[resultaxis,
    xmin=0,xmax=5.2,ymin=0.55,ymax=1.08,
    xtick={0,1,2,3,4,5},ytick={0.6,0.7,0.8,0.9,1.0},
    xlabel={Iteration},ylabel={Normalized area},
    every axis plot/.append style={line width=0.9pt},
    legend cell align=left,
]

\addplot[resultorange,mark=o,mark size=1.7pt,mark options={fill=white}] coordinates {
    (0,1.000)(1,0.744)(2,0.862)(3,0.626)(4,0.687)(5,0.800)
};
\addlegendentry{Cand.}
\addplot[resultblue,dashed,mark=*,mark size=1.2pt] coordinates {
    (0,1.000)(1,0.744)(2,0.744)(3,0.626)(4,0.626)(5,0.626)
};
\addlegendentry{Best}
\addplot[black!30,densely dashed,forget plot] coordinates {(0,1)(5.2,1)};
\end{axis}
\end{tikzpicture}
\caption{Optimization.}
\label{fig:ppa_traj}
\end{subfigure}\hfill
\begin{subfigure}[t]{0.24\textwidth}
\centering
\begin{tikzpicture}
\begin{axis}[resultaxis,
    xbar=1pt,bar width=3pt,
    xmin=0,xmax=40,ymin=0.5,ymax=6.5,
    xtick={0,20,40},ytick={1,2,3,4,5,6},
    yticklabels={Comb.,Case-z,Shift,Add/Sub,Mux,FSM},
    xlabel={Reduction (\%)},
    ymajorgrids=false,xmajorgrids=true,
    legend image code/.code={\draw[#1] (0cm,-0.05cm) rectangle (0.18cm,0.05cm);},
]
\addplot[fill=resultblue,draw=none] coordinates {
    (9.9,1) (10.4,2) (11.4,3) (17.4,4) (27.3,5) (35.2,6)
};
\addlegendentry{Area}
\addplot[fill=resultteal,draw=none] coordinates {
    (6.4,1) (2.0,2) (7.5,3) (8.9,4) (14.5,5) (30.3,6)
};
\addlegendentry{Power}
\end{axis}
\end{tikzpicture}
\caption{Area--power gains.}
\label{fig:ppa_opt}
\end{subfigure}\hfill
\begin{subfigure}[t]{0.24\textwidth}
\centering
\begin{tikzpicture}
\begin{axis}[resultaxis,
    xmin=40,xmax=105,ymin=0.4,ymax=4.7,
    xtick={50,75,100},ytick={1,2,3,4},
    yticklabels={Yosys,Proof,Sim.,Compile},
    xlabel={Pass rate (\%)},
    ymajorgrids=false,xmajorgrids=true,
    clip=false,
]
\addplot[resultgray!65,line width=1.4pt,forget plot] coordinates {(76.7,4)(100,4)};
\addplot[resultgray!65,line width=1.4pt,forget plot] coordinates {(50,3)(70,3)};
\addplot[resultgray!65,line width=1.4pt,forget plot] coordinates {(53.3,2)(63.3,2)};
\addplot[resultgray!65,line width=1.4pt,forget plot] coordinates {(69.6,1)(73.3,1)};
\addplot[only marks,mark=*,mark size=1.2pt,resultorange,mark options={fill=resultorange}] coordinates {(76.7,4)(50,3)(53.3,2)(69.6,1)};
\addlegendentry{1-shot}
\addplot[only marks,mark=*,mark size=1.2pt,resultblue] coordinates {(100,4)(70,3)(63.3,2)(73.3,1)};
\addlegendentry{Step}
\node[anchor=north east,font=\tiny,text=resultorange,inner sep=1pt] at (axis cs:76.7,4) {76.7};
\node[anchor=north east,font=\tiny,text=resultorange,inner sep=1pt] at (axis cs:50,3) {50.0};
\node[anchor=north east,font=\tiny,text=resultorange,inner sep=1pt] at (axis cs:53.3,2) {53.3};
\node[anchor=north east,font=\tiny,text=resultorange,inner sep=1pt] at (axis cs:69.6,1) {69.6};
\node[anchor=south east,font=\tiny,text=resultblue,inner sep=0pt] at (axis cs:100,4) {100};
\node[anchor=south west,font=\tiny,text=resultblue,inner sep=1pt] at (axis cs:70,3) {70.0};
\node[anchor=south west,font=\tiny,text=resultblue,inner sep=1pt] at (axis cs:63.3,2) {63.3};
\node[anchor=south west,font=\tiny,text=resultblue,inner sep=1pt] at (axis cs:73.3,1) {73.3};
\end{axis}
\end{tikzpicture}
\caption{Proof generation.}
\label{fig:proof_ablation}
\end{subfigure}\hfill
\begin{subfigure}[t]{0.24\textwidth}
\centering
\makebox[\linewidth][r]{%
\begin{tikzpicture}
\begin{axis}[resultaxis,
    xmin=0,xmax=5.2,ymin=-0.50,ymax=0.48,
    xtick={0,1,2,3,4,5},
    ytick={-0.4,0,0.4},
    xlabel={Iteration},ylabel={Slack (ns)},
    ylabel style={font=\scriptsize},
    ylabel shift=-7pt,
    xmajorgrids=false,
    every axis plot/.append style={line width=0.9pt},
    legend columns=4,
    legend style={column sep=1pt},
    legend image post style={xscale=0.4},
    legend cell align=left,
]
\addplot[black!45,densely dashed] coordinates {(0,-0.37)(5.2,-0.37)};
\addlegendentry{WNS}
\addplot[resultorange,mark=triangle*,mark size=1.3pt,mark options={fill=resultorange}] coordinates {
    (0,-0.37)(1,0.14)(2,0.09)(3,0.18)(4,0.11)(5,0.12)
};
\addlegendentry{A}
\addplot[resultblue,mark=*,mark size=1.3pt] coordinates {
    (0,-0.37)(1,0.05)(2,0.35)(3,0.20)(4,0.35)(5,0.35)
};
\addlegendentry{T}
\addplot[resultteal,mark=square*,mark size=1.3pt,mark options={fill=resultteal}] coordinates {
    (0,-0.37)(1,0.11)(2,0.31)(3,0.29)(4,0.29)(5,0.29)
};
\addlegendentry{P}
\addplot[black!30,forget plot] coordinates {(0,0)(5.2,0)};
\node[font=\tiny,anchor=west,inner sep=1pt] at (axis cs:0.12,-0.37) {$-0.37$};
\end{axis}
\end{tikzpicture}\hspace*{3pt}}

\caption{Tight-clock slack.}
\label{fig:wns_slack}
\end{subfigure}
\caption{(a)~Measured Yosys synthesis-area trajectory: orange is the candidate at each iteration; blue is the accepted best-so-far (running minimum). Iterations~2, 4, and~5 are rejected worse candidates; the best area first appears at iteration~3 and is held through iteration~5. The gray dashed line marks the seed area ($=1$). (b)~Area and power reductions for the six simulation-passing examples in Table~\ref{tab:arch_explore_reductions}. (c)~One-shot vs.\ stepwise proof generation on a 30-problem VerilogEval pilot. (d)~Post-route setup slack over $K{=}5$ iterations under a tight clock. The dashed line is the seed WNS ($-0.37\,\mathrm{ns}$); orange/blue/teal are area, timing, and power feedback. The gray line at 0 is timing closure.}
\label{fig:ppa_and_proof}
\end{figure}
\Needspace{23\baselineskip}
\begin{wrapfigure}[21]{r}{0.42\textwidth}
    \raggedleft
    \includegraphics[width=\linewidth]{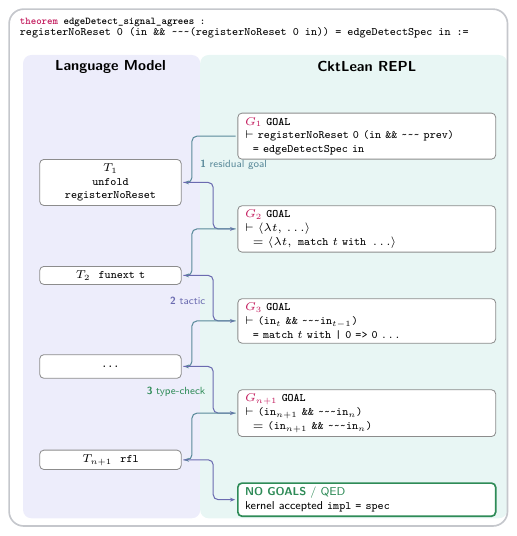}
    \captionsetup{justification=justified,singlelinecheck=false,skip=4pt}
    \caption{Stepwise proof of the rising-edge detector equivalence lemma: the registered implementation matches its specification for every input stream. Teal: goals (REPL $\to$ model); violet: tactics (model $\to$ REPL).}
    \label{fig:stepwise_edgedetect}
\end{wrapfigure}

\paragraph{Functional Correctness.}
\textsc{CktFormalizer} exceeds RTLCoder and CodeV in RTL simulation on all
four benchmarks, but does not consistently lead direct-generation baselines:
Verilog Agent leads on VerilogEval and ResBench, while ICL leads on RTLLM
and CVDP (Table~\ref{tab:main_results}). Among RTL-simulation-passing designs,
\textsc{CktFormalizer} achieves 88.2\% gate-level simulation pass rates after
both synthesis and P\&R on RTLLM, and 100.0\% on ResBench, exceeding ICL
with SystemVerilog and Verilog Agent at both stages. Its GLS-P\&R rate matches Verilog Agent
on VerilogEval (94.6\%) but falls below both baselines on CVDP
(27.9\% vs.\ 33.3\%). These results demonstrate stronger backend functional robustness
on RTLLM and ResBench relative to the direct SystemVerilog baselines.
\emph{With PD Pass rates of 95.4\%--100.0\% among RTL-simulation-passing
designs, \textsc{CktFormalizer} demonstrates that hardware generated through
\textsc{CktLean}'s typed abstractions can translate into physically realizable
implementations.}
\par
\begin{tcolorbox}[width=\linewidth,colback=gray!10,colframe=gray!40,boxrule=0.5pt,
    arc=2pt,left=7pt,right=7pt,top=5pt,bottom=5pt,
    before skip=7pt,after skip=7pt]
\textbf{Findings.} \emph{\textsc{CktLean}'s typed abstractions enable hardware
generation that extends beyond simulation to physical implementation.}
\end{tcolorbox}
\vspace{-2mm}

\subsection[Hardware Optimization and Verification]{\mbox{Hardware Optimization and Verification}}

\label{sec:optimization_verification}

\vspace{-2mm}
\paragraph{Equivalence Proof Generation.}
\label{sec:proof_ablation}
\emph{As a step toward correctness-preserving PPA optimization, we explore whether stepwise Lean feedback enables LLMs to prove implementation--specification equivalence.}
We run a 30-problem VerilogEval pilot to isolate the effect of proof-generation strategy. Figure~\ref{fig:ppa_and_proof}(c) compares the two modes.
(i)~\emph{One-shot.} The model writes the implementation, specification, and Lean proof in a single attempt without intermediate proof-goal feedback.
(ii)~\emph{Stepwise.} The model can query Lean proof states and iteratively refine the theorem.
A proof counts as complete only if Lean's kernel accepts the final theorem without \texttt{sorry} or \texttt{admit}.
\par

Stepwise proof development improves all metrics, raising compile pass from 77\% to 100\%, simulation from 50\% to 70\%, kernel-accepted proof completion from 53.3\% (16/30) to 63.3\% (19/30), and Yosys equivalence from 70\% to 73\%. These completed proofs establish source-level formal correctness relative to their stated Lean specifications.

Figure~\ref{fig:stepwise_edgedetect} illustrates the proof for a registered
rising-edge detector whose output must match its specification for every input
stream. The model unfolds definitions and refines tactics using REPL feedback until
Lean's kernel accepts the proof that the implementation and specification
produce identical outputs for every input stream.
\begin{tcolorbox}[width=\linewidth,colback=gray!10,colframe=gray!40,boxrule=0.5pt,
    arc=2pt,left=7pt,right=7pt,top=5pt,bottom=5pt,
    before skip=7pt,after skip=7pt]
\textbf{Findings.} \emph{By exposing intermediate proof states beyond whole-file compiler feedback, the \textsc{CktLean} REPL improves proof completion through stepwise refinement.}
\end{tcolorbox}
\vspace{-2mm}
\paragraph{Architecture Exploration.}
\label{sec:arch_results}
We evaluate architecture exploration on the 63-problem VerilogEval subset (\texttt{Prob035}--\texttt{Prob156}, the more complex half of the 156-problem benchmark, excluding simple wiring and gate tasks), generating $C{=}4$ candidates per design with different micro-architectural choices.
(i)~\emph{Coverage.} Of 63 problems, 27 produced multiple successfully synthesized candidates, enabling cross-candidate comparison.
(ii)~\emph{Overall gains.} Among these 27 designs, 10 (37.0\%) achieved synthesis-level area reductions exceeding 5\%.
(iii)~\emph{Illustrative examples.} Table~\ref{tab:arch_explore_reductions} presents six examples with simulation-passing candidates, with area reductions of 9.9\%--35.2\% and post-route power reductions of 2.0\%--30.3\%. Appendix~\ref{app:arch_explore_data} explains the design labels. Figure~\ref{fig:ppa_and_proof}(b) plots the area and power reductions for these six examples.

\par
\begin{tcolorbox}[width=\textwidth,colback=gray!10,colframe=gray!40,boxrule=0.5pt,
    arc=2pt,left=7pt,right=7pt,top=5pt,bottom=5pt,
    before skip=7pt,after skip=7pt]
\textbf{Findings.} \emph{Synthesis feedback guides PPA optimization, while equivalence proofs can support more aggressive changes by verifying source-level behavioral preservation.}
\end{tcolorbox}

\Needspace{14\baselineskip}

\begin{wraptable}[15]{r}{0.40\textwidth}

\raggedleft
\captionsetup{font=normalsize,skip=3pt,justification=justified,singlelinecheck=false}
\caption{Area, cell-count, and power reductions from architecture exploration.}
\label{tab:arch_explore_reductions}
\small
\setlength{\tabcolsep}{2pt}
\renewcommand{\arraystretch}{1.22}
\begin{tabular*}{\linewidth}{@{\extracolsep{\fill}}lrrr@{}}
\toprule[0.6pt]
\multirow{2}{*}{\textbf{Design}} & \multicolumn{3}{c}{\textbf{Reduction (\%)}} \\
\cmidrule(l){2-4}
 & Area & Cells & Power \\
\midrule[0.35pt]
FSM (Lemmings) & \textbf{35.2} & 35.0 & \textbf{30.3} \\
9-to-1 Mux & 27.3 & \textbf{45.9} & 14.5 \\
Add/Sub Unit & 17.4 & 26.2 & 8.9 \\
FSM Shift Reg. & 11.4 & 0.0 & 7.5 \\
Case-z Logic & 10.4 & 0.0 & 2.0 \\
Comb. Logic & 9.9 & 14.8 & 6.4 \\
\bottomrule[0.6pt]
\end{tabular*}
\vspace{-2mm}
\end{wraptable}

\vspace{-2mm}
\paragraph{PPA Optimization Loop.}
\label{sec:ppa_opt_results}
We run $K{=}5$ PPA optimization iterations on the 63-problem VerilogEval subset, using the simulation-passing semantic \textsc{CktFormalizer} seeds. 54 designs enter the loop (9 seeds are not \texttt{sim\_pass}); 31 change Lean and reduce Yosys area, of which 28 exceed $\tau{=}5\%$. Figure~\ref{fig:ppa_and_proof}(a) plots a measured candidate vs.\ best-so-far trajectory.
(i)~\emph{Effectiveness.} The largest Yosys-area reduction is $-$58.6\% ($196\to81\,\mu\mathrm{m}^{2}$). This loop uses synthesis-level area and cell count.
(ii)~\emph{Non-monotonic trajectories.} Optimization trajectories are non-monotonic---Lean's type system rejects unsafe transformations, limiting the search space but preventing optimization-induced bugs.
(iii)~\emph{Comparison with architecture exploration.} Architecture exploration generates and compares alternative microarchitectures (Table~\ref{tab:arch_explore_reductions} and Figure~\ref{fig:ppa_and_proof}(b)), whereas this loop iteratively modifies a selected design.
(iv)~\emph{Tight-clock P\&R feedback.} A separate pass feeds post-route setup slack and power to the agent. Figure~\ref{fig:ppa_and_proof}(d) shows one tight-clock case: timing and power arms close WNS (setup slack from $-0.37$ to $+0.35$ and $+0.31\,\mathrm{ns}$). In the area-focused optimization run, the selected design's setup slack remains at the initial value of $-0.37\,\mathrm{ns}$.
\emph{This example shows that post-route feedback can guide optimization toward timing closure, which area-focused optimization alone does not ensure.}

\FloatBarrier
\vspace{-3mm}

\section{Conclusion}
\vspace{-3mm}
\label{sec:conclusion}

We presented \textsc{CktFormalizer}, an agent framework built on our core
contribution, \textsc{CktLean}, which combines typed hardware representations,
compilation, and a persistent Lean REPL. Across four benchmarks, it generates
hardware that passes structural checks and supports physical implementation.
Proof-state feedback improves kernel-checked equivalence proof completion,
while hardware evaluation guides architecture exploration and PPA optimization.
Typed infrastructure and feedback thus support LLM-driven circuit generation,
optimization, and verification.

\FloatBarrier
\clearpage
\subsection*{AI use statement}

Generative AI is both the subject of study and an implementation component of
the evaluated system: language models generate candidate Lean or SystemVerilog
artifacts under the workflows described in Section~\ref{sec:method}, particularly
Section~\ref{sec:eval_pipeline}. These artifacts are checked
with the compiler, simulation, proof, synthesis, equivalence, and physical-design
stages reported in the paper. An AI coding assistant was additionally used to
migrate the manuscript source to the ICLR template and check formatting. The
author reviewed the AI-assisted manuscript changes and takes responsibility for
the final text, claims, citations, and artifacts.

\bibliographystyle{iclr2027_conference}
\bibliography{custom}

\clearpage
\appendix
\section{\textsc{Sparkle} Circuit-Level Proofs Across Versions}
\label{app:sparkle_proof_examples}

\textsc{Sparkle}'s verification surface differs across releases. Our experimental
pipeline pins v0.1.0~\citep{hashimoto2026sparkle}. That release provides typed
\texttt{Signal} construction, cycle-level evaluation, and SystemVerilog
extraction, and includes Lean lemmas over pure \texttt{BitVec} and ISA models;
it does not yet contain a dedicated bridge from synthesizable \texttt{Signal}
terms to those specifications or a theorem for unfolding feedback loops. Later
revisions add \texttt{SignalDSLProps}, \texttt{Equivalence},
\texttt{Temporal}, and \texttt{LoopProps}
modules~\citep{verilean2026sparkleverification}. The two examples below use the
current, post-v0.1.0 interfaces to illustrate capabilities now available in
\textsc{Sparkle}; they should not be read as capabilities exercised by the v0.1.0
experimental runs. Both examples establish source-level properties only and do
not prove semantic preservation of SystemVerilog extraction.

\subsection{Connecting a Signal Circuit to a Pure Specification}
\label{app:sparkle_signal_bridge}

Listing~\ref{lst:sparkle_signal_bridge} separates a pure functional
specification from the synthesizable \texttt{Signal} implementation. The
theorem quantifies over arbitrary input streams and every cycle \texttt{t}; evaluating the circuit at
\texttt{t} yields the same value as applying the pure specification to the
inputs at \texttt{t}. Lean accepts the proof by definitional reduction, so its
kernel checks the connection to the actual \texttt{Signal} term rather than to
an unrelated test model. This deliberately minimal example demonstrates the
semantic bridge, not a complex functional property.

\begin{lstlisting}[style=lean4,caption={A combinational \texttt{Signal} circuit and its pure functional specification.},label={lst:sparkle_signal_bridge}]
import Sparkle

open Sparkle.Core.Domain
open Sparkle.Core.Signal

namespace AppendixProofs

def addCircuit
    (a b : Signal defaultDomain (BitVec 8)) :
    Signal defaultDomain (BitVec 8) :=
  a + b

def addSpec (a b : BitVec 8) : BitVec 8 :=
  a + b

theorem addCircuit_correct
    (a b : Signal defaultDomain (BitVec 8)) (t : Nat) :
    (addCircuit a b).val t = addSpec (a.val t) (b.val t) := by
  rfl

#synthesizeVerilog addCircuit

end AppendixProofs
\end{lstlisting}

\subsection{Reasoning About a Registered Feedback Loop}
\label{app:sparkle_feedback_proof}

Listing~\ref{lst:sparkle_feedback_proof} is sequential. The implementation
closes a feedback path through \texttt{Signal.register}, while
\texttt{counterPure} is an ordinary recursive function over cycles. The proof
first establishes strict causality, then applies \textsc{Sparkle}'s kernel-checked
\texttt{loop\_unfold} fixpoint theorem explicitly. The subsequent
\texttt{loop\_iterate} proof connects the feedback circuit to the pure recurrence
at every cycle, rather than only over a bounded simulation trace.

\begin{lstlisting}[style=lean4,literate={·}{{$\cdot$}}1,caption={An all-cycle refinement proof for a registered feedback counter.},label={lst:sparkle_feedback_proof}]
import Sparkle
import Sparkle.Verification.LoopProps

open Sparkle.Core.Domain
open Sparkle.Core.Signal
open Sparkle.Verification.LoopProps

namespace AppendixProofs

def counterBody (s : Signal defaultDomain (BitVec 8)) :
    Signal defaultDomain (BitVec 8) :=
  Signal.register 0#8 (s + 1#8)

def counterLoop : Signal defaultDomain (BitVec 8) :=
  Signal.loop counterBody

def counterPure : Nat -> BitVec 8
  | 0 => 0#8
  | t + 1 => counterPure t + 1#8

theorem counterBody_strict : StrictlyCausal counterBody := by
  apply strictlyCausal_of_step
    (c0 := 0#8)
    (next := fun _ x => x + 1#8)
  · intro s
    rfl
  · intro s t
    simp [counterBody]

theorem counterLoop_unfold :
    counterLoop = counterBody counterLoop := by
  exact loop_unfold counterBody counterBody_strict

theorem counterLoop_correct (t : Nat) :
    counterLoop.val t = counterPure t := by
  apply loop_iterate
    (f := counterBody)
    (c0 := 0#8)
    (next := fun _ x => x + 1#8)
    (st := counterPure)
  · intro s
    rfl
  · intro s t
    simp [counterBody]
  · rfl
  · intro t
    rfl

theorem counterLoop_counts (t : Nat) :
    counterLoop.val t = BitVec.ofNat 8 t := by
  rw [counterLoop_correct]
  induction t with
  | zero => rfl
  | succ u ih =>
      rw [counterPure, ih]
      bv_omega

#synthesizeVerilog counterLoop

end AppendixProofs
\end{lstlisting}

\section{Experimental Setup Details}
\label{app:experimental_setup}
\label{app:tool_report}

This appendix records the \textsc{Sparkle} main-flow protocol used for
Table~\ref{tab:main_results}: benchmark inputs, model and action-budget
settings, the agent tool interface, and the evaluator used after each
generation or compile-only repair session. Architecture exploration,
in-the-loop PPA, simulation-mismatch repair, and the extra skill-pack
prompt were disabled for these runs.

\subsection{Benchmarks and Task Inputs}
\label{app:benchmarks}

We evaluate RTL generation on four benchmarks.
VerilogEval~\citep{liu2023verilogeval} contains 156 HDLBits-derived
problems spanning combinational logic, sequential circuits, counters,
shift registers, and finite-state machines.
RTLLM~\citep{lu2024rtllm} contains 50 natural-language RTL tasks covering
arithmetic, memory, control, and miscellaneous modules.
ResBench~\citep{guo2025resbench} contains 56 arithmetic,
signal-processing, control, and datapath tasks.
CVDP~\citep{pinckney2025cvdp} uses the official 168-problem
non-agentic, no-commercial code-generation split, not the full
783-problem release. That split covers single-turn RTL generation
without commercial EDA tools or restricted IP; 156 tasks are ordinary
fixed-width designs and 12 retain native parameters.

The coding agent sees only the public specification. On CVDP this is the
official prompt and context together with the evaluation top-module name.
Hidden testbenches, gold RTL bodies, and simulator traces are withheld.
VerilogEval, RTLLM, and ResBench already expose public natural-language
specifications and use the same agent loop. Reference RTL may be used by
the evaluator for wrappers; it is not an agent input. The \textsc{Sparkle} agent
writes one Lean HDL file per problem. The evaluator extracts
SystemVerilog and scores it with the hidden harness. Direct-SystemVerilog
baselines use the same backbone model but emit Verilog rather than Lean.

The boxes below reproduce the generation templates used for
Table~\ref{tab:main_results}. Placeholders stand for the public
specification and required module name. ICL is a single call and does not insert worked solutions: its
only in-context fragment is the required port header. \textsc{CktFormalizer}
is the method that uses in-context examples: the workspace mounts 13 generic
Lean idiom files (combinational, sequential, and FSM sketches, including
\texttt{RTLIdioms.lean}), and the compact skill allows at most two reads
(three on CVDP). These files are style references, not evaluation gold;
their full contents appear in
Appendix~\ref{app:idiom_library}. 

\begin{tcolorbox}[casebox,title={RTLCoder (pass@1)}]
\begin{lstlisting}[style=logexcerpt]
Please act as a professional verilog designer.

<public natural-language specification>

module TopModule(
  input  clk,
  input  reset,
  output [7:0] out
);
\end{lstlisting}
\end{tcolorbox}

\begin{tcolorbox}[casebox,title={CodeV (pass@1)}]
\begin{lstlisting}[style=logexcerpt]
<Verilog>
Generate synthesizable SystemVerilog for the following hardware task.
Output only one complete module, from module declaration to endmodule.
The module name must be TopModule.
Keep the exact external port interface shown below.

Natural language specification:
<public natural-language specification>

Required module interface:
module TopModule(
  input  clk,
  input  reset,
  output [7:0] out
);

SystemVerilog implementation:
\end{lstlisting}
\end{tcolorbox}

\begin{tcolorbox}[casebox,title={ICL (one-shot gpt-5.6-sol)}]
\textbf{System.}
\begin{lstlisting}[style=logexcerpt]
You are an expert hardware designer. Generate synthesizable SystemVerilog code.

RULES:
1. The module MUST be named `TopModule` with EXACTLY the ports specified.
2. Output ONLY the SystemVerilog module code (module ... endmodule). No explanation.
3. Use SystemVerilog-2012 syntax compatible with Icarus Verilog (-g2012).
4. For sequential logic, use `always_ff @(posedge clk)` when appropriate.
5. For combinational logic, use `assign` or `always_comb`.
\end{lstlisting}
\textbf{User (in-context port header; gold body withheld).}
\begin{lstlisting}[style=logexcerpt]
Specification:
<public natural-language specification>

Reference port interface (your module MUST be named TopModule with the same ports):
```
module RefModule (
  input  clk,
  input  reset,
  output [7:0] out
);
```

Generate the complete `TopModule` implementation:
\end{lstlisting}
\end{tcolorbox}

\begin{tcolorbox}[casebox,title={Harness (Archon compile-fix, 100 turns)}]
\begin{lstlisting}[style=logexcerpt]
You are an expert hardware designer using an Archon-style agent harness
to implement RTL directly in SystemVerilog.

## Goal
Generate one synthesizable SystemVerilog module that matches the
benchmark contract and passes the external compile/simulation harness.

## Output File
- Write the candidate to cktarchon_work/<id>/candidate.sv.
- The file must contain complete SystemVerilog with module ... endmodule.
- Do not write Lean or Sparkle.

## RTL Rules
- Use SystemVerilog-2012 syntax compatible with Icarus Verilog/cocotb.
- Match the required top-module name, ports, widths, and clock/reset.
- Avoid unsynthesizable constructs in the DUT.

## Workflow
1. Read the problem and benchmark interface contract.
2. Write a complete candidate.
3. If compile-only feedback is provided, repair the same file.
4. Stop after the candidate file is written.

## Problem
- Required top module: `TopModule`
- Natural-language specification: <public specification>
\end{lstlisting}
\end{tcolorbox}

\begin{tcolorbox}[casebox,title={\textsc{CktFormalizer} compact skill}]
\begin{lstlisting}[style=logexcerpt]
You are an expert hardware engineer translating natural-language RTL
specifications into Sparkle HDL, a Lean 4 hardware DSL.

## File Template
import Sparkle
import Sparkle.Compiler.Elab
open Sparkle.Core.Signal
def <target> {dom : DomainConfig} (<inputs>) : <output_type> :=
  <implementation>
#synthesizeVerilog <target>

## Workflow
1. Read at most two small examples, preferably Benchmark/RTLIdioms.lean
   plus one similar Benchmark/*.lean.
2. Write a complete candidate to Generated/<id>.lean.
3. Run the harness Lean check and fix compiler/extraction errors.
4. Stop after the Lean check reports success.
\end{lstlisting}
\end{tcolorbox}

\subsection{Model and Hyperparameters}
\label{app:model_tool_setup}

All \textsc{Sparkle} main-flow runs use a Codex coding agent with
\texttt{gpt-5.6-sol}, reasoning effort ultra, the Responses API, and
16{,}384 maximum output tokens, in an isolated workspace-write sandbox.
A turn is one Codex action. Generation and compile-only repair share 100
actions: the initial Lean candidate is capped at 40, and leftover actions
are used for repair in at most 9 rounds of at most 10 actions. The
evaluator may retry infrastructure failures without spending model turns.

\paragraph{\textsc{Sparkle} Version.}
All reported experiments pin \textsc{Sparkle} v0.1.0 (commit
\texttt{13458ca}). Upstream v0.1.0 does not contain the imperative circuit
notation used in our experimental source; the experimental stack provides a
local \texttt{Signal.circuit do} front end that lowers to
\texttt{Signal.loop} and \texttt{Signal.register}. \textsc{Sparkle} subsequently added
the qualified \texttt{Signal.circuit do} form in commit \texttt{d22cb7d} and
later replaced it with the unqualified \texttt{circuit do} form in commit
\texttt{104f2cc}~\citep{verilean2026sparklecircuitdsl}. The circuit-level proof
interfaces illustrated in
Appendix~\ref{app:sparkle_proof_examples} use the later commit
\texttt{fa0e48e} and document \textsc{Sparkle}'s subsequent evolution; these interfaces
were not exercised in the reported experimental runs.

\paragraph{System prompt and idioms.}
Main runs use a compact \textsc{Sparkle} skill plus harness rules: import
\texttt{Sparkle} and \texttt{Sparkle.Compiler.Elab}, write one Lean
candidate per problem, run the prescribed Lean check until SystemVerilog
is emitted, and stop. The agent does not run simulation, cocotb, Yosys,
or place-and-route. The sandbox mounts 13 generic idiom files, including
\texttt{RTLIdioms.lean} and small combinational, sequential, and FSM
sketches. These files are not evaluation gold; they are listed in
Appendix~\ref{app:idiom_library}. The prompt allows at most
two reads, preferably \texttt{RTLIdioms.lean} and one similar file (at
most three on CVDP).

\paragraph{Compile-only repair.}
Repair uses only pre-simulation failures: Lean elaboration, missing
SystemVerilog, parameter-contract errors, and \texttt{iverilog}
lint or elaboration. On CVDP, a public port check may trigger at most
one structural repair. Hidden simulation failures are not fed back.
Post-hoc PPA on sky130hd (10\,ns clock) does not call the model.
Table~\ref{tab:hyperparams} lists the remaining knobs.

\begin{table}[ht]
\centering
\caption{\textsc{Sparkle} main-flow hyperparameters.}
\label{tab:hyperparams}
\label{tab:system_prompt}
\small
\begin{tabular}{ll}
\toprule
\textbf{Setting} & \textbf{Value} \\
\midrule
Model / effort / max tokens & \texttt{gpt-5.6-sol}, ultra, 16{,}384 \\
Harness & Codex, workspace-write, Responses API \\
Shared action budget & 100 (generation cap 40) \\
Compile-only repair & at most 9 rounds $\times$ 10 actions \\
CVDP structural repair & at most one public port check \\
Feedback / hidden simulation & compile-only / withheld \\
CVDP / other dataset specs & public-spec-v2 / public NL specs \\
Idiom files & 13 mounted; read at most 2 (3 on CVDP) \\
\bottomrule
\end{tabular}
\end{table}

\subsection{Component Ablation on VerilogEval Subset}
\label{app:component_ablation}

Table~\ref{tab:component_ablation_verilogeval50} reports a component ablation
on a fixed 50-problem VerilogEval subset sampled uniformly at random without
replacement using seed 20260505. The Full row is filtered from an existing
backend-enabled 156-problem VerilogEval run, and the ablation rows are
recomputed from per-problem result records on the same subset. All metrics use
the 50 sampled problems as the denominator. Each ablation row reports the
result after removing one component from the full system. Removing
\emph{Iterative Repair} disables the multi-turn repair loop after the initial
Lean HDL generation, so the model cannot use compiler or evaluator feedback to
edit the design. Removing \emph{Lean REPL Feedback} keeps the outer generation
loop but removes interactive Lean REPL state and proof-goal feedback, leaving
the agent to rely on file edits and batch compiler diagnostics. Removing
\emph{Few-shot Examples} removes the verified Lean HDL examples from the prompt
context and blocks access to the benchmark example library during generation.

\begin{table}[t]
\centering
\caption{Component ablation on the fixed 50-problem VerilogEval subset. Metrics are recomputed from per-problem JSONL records and reported independently; skipped or missing downstream checks count as failures under the same 50-problem denominator.}
\label{tab:component_ablation_verilogeval50}
\small
\resizebox{\columnwidth}{!}{%
\begin{tabular}{lcccccccc}
\toprule
\textbf{Removed Component} & \textbf{Compile} & \textbf{Sim.} & \textbf{Synth} & \textbf{P\&R} & \textbf{DRC} & \textbf{LVS} & \textbf{GLS-Synth} & \textbf{GLS-P\&R} \\
\midrule
None (Full) & 50/50 (100.0\%) & 35/50 (70.0\%) & 50/50 (100.0\%) & 47/50 (94.0\%) & 47/50 (94.0\%) & 47/50 (94.0\%) & 34/50 (68.0\%) & 32/50 (64.0\%) \\
Iterative Repair & 44/50 (88.0\%) & 22/50 (44.0\%) & 32/50 (64.0\%) & 31/50 (62.0\%) & 31/50 (62.0\%) & 31/50 (62.0\%) & 22/50 (44.0\%) & 21/50 (42.0\%) \\
Lean REPL Feedback & 50/50 (100.0\%) & 35/50 (70.0\%) & 50/50 (100.0\%) & 47/50 (94.0\%) & 47/50 (94.0\%) & 47/50 (94.0\%) & 34/50 (68.0\%) & 33/50 (66.0\%) \\
Few-shot Examples & 49/50 (98.0\%) & 34/50 (68.0\%) & 49/50 (98.0\%) & 47/50 (94.0\%) & 47/50 (94.0\%) & 47/50 (94.0\%) & 34/50 (68.0\%) & 32/50 (64.0\%) \\
\bottomrule
\end{tabular}%
}
\end{table}

\FloatBarrier

\subsection{Agent Tools and Evaluation Pipeline}
\label{app:agent_callable_tools}
\label{app:eval_pipeline}

The Codex agent may read mounted \textsc{Sparkle} sources and the idiom library,
edit the per-problem Lean file, search the repository, and run a Lean
check until SystemVerilog is emitted. After each generation or repair
session the outer evaluator, which holds
the hidden harness, runs the same ordered checks. Lean elaboration (REPL
or \texttt{lake build}) executes
\texttt{\#synthesizeVerilog} or
\texttt{\#synthesizeParameterizedVerilog}. The harness extracts
SystemVerilog, optionally wraps ports onto the benchmark interface, and
runs \texttt{iverilog} lint (\texttt{-t null -g2012}). Native CVDP tasks
elaborate every public parameter case. CVDP additionally checks public
port names, directions, and numeric widths. Functional simulation then
uses the hidden testbench: Icarus Verilog and \texttt{vvp} on VerilogEval,
RTLLM, and ResBench; cocotb on CVDP. Compile-only repair may use
diagnostics from stages before simulation; candidate selection uses
compile/extract/lint and public-structure progress, not simulation scores.

Designs that reach \texttt{sim\_pass} may be sent to a separate OpenROAD
Flow Scripts backend on \texttt{sky130hd} (Yosys, place-and-route, DRC,
LVS, optional gate-level simulation). That backend does not call the
model and does not change the simulation-pass numbers in the main table.

\subsection{EDA Tools}
\label{app:eda_tools}

Table~\ref{tab:eda_tools} lists tools used after a Lean candidate exists.
Lean and Icarus participate in the agent-visible compile path; Yosys,
OpenROAD, and KLayout run only in the post-hoc backend. PPA metrics and
Yosys \texttt{equiv\_*} checks are not returned to the \textsc{Sparkle} coding
agent in the main flow.

\begin{table}[t]
\centering
\caption{EDA tools used by the automated evaluation and backend implementation
flow.}
\label{tab:eda_tools}
\small
\begin{tabular}{p{0.22\linewidth}p{0.24\linewidth}p{0.44\linewidth}}
\toprule
\textbf{Tool} & \textbf{Pipeline Stage} & \textbf{Role} \\
\midrule
\texttt{Lean 4} & Source compilation &
Type-checks Lean HDL and runs the \texttt{\#synthesizeVerilog} metaprogram
to emit synthesizable SystemVerilog. \\
\texttt{Icarus Verilog} & RTL lint and simulation &
Runs \texttt{iverilog -t null -g2012} for syntax checking and
\texttt{iverilog -g2012} for testbench compilation. \\
\texttt{vvp} & RTL and gate-level simulation &
Executes the simulator output produced by Icarus Verilog and records testbench
pass/fail status or mismatch counts. \\
\texttt{Yosys} & Post-hoc logic synthesis &
Performs synthesis inside the OpenROAD Flow Scripts backend. Optional
\texttt{equiv\_*} RTL checks are not part of the \textsc{Sparkle} generation loop. \\
\texttt{OpenROAD Flow} & Backend orchestration &
A set of scripts that orchestrate the EDA backend flow (synthesis, place-and-route, DRC, LVS) inside a Docker container, enabling reproducible physical implementation from RTL to GDSII. \\
\texttt{OpenROAD}/\texttt{OpenSTA} & Physical implementation and timing &
OpenROAD is an open-source digital design implementation tool; \texttt{OpenSTA} is its static timing analysis engine. Together they perform floorplanning, placement, clock-tree synthesis, routing, finishing, and static timing analysis (STA); the evaluator extracts WNS, power, and PPA metrics from OpenROAD reports. \\
\texttt{KLayout} & Physical verification &
Runs DRC and LVS checks through the OpenROAD Flow signoff targets. \\
\texttt{Docker} & Reproducible execution &
Runs the backend flow in the \texttt{openroad/orfs:latest} container with
mounted result, log, and report directories. \\
\texttt{SkyWater sky130hd} & Technology target &
Provides the \texttt{sky130\_fd\_sc\_hd} standard-cell library and PDK
configuration used for synthesis, place-and-route, timing, DRC, and LVS. \\
\texttt{volare} & Cell-model management &
Fetches SkyWater standard-cell simulation models when gate-level simulation
requires local cell libraries. \\
\texttt{netlistsvg} & Optional visualization &
Renders Yosys-generated JSON netlists into schematic SVGs for inspection. \\
\bottomrule
\end{tabular}
\end{table}
\FloatBarrier

\subsection{Error Analysis: Case Study}
\label{app:error_analysis}

We manually inspect the 31 VerilogEval designs that compile in Lean but do not
pass RTL simulation in the main run. The boxes below give representative cases
with the archived evaluator signal and the corresponding testbench evidence.

\begin{tcolorbox}[casebox,title={Case 1: Reset timing -- \texttt{Prob047\_dff8ar}}]
\textbf{Prompt.} Eight D flip-flops with active-high asynchronous reset.
\begin{lstlisting}[style=logexcerpt]
log: compile=pass, lint=pass, sim=fail, synth/pnr/gds=pass
     44 mismatches in 436 samples
tb:  reset_test(1);
     Hint: reset should be asynchronous/synchronous.
\end{lstlisting}
\end{tcolorbox}

\begin{tcolorbox}[casebox,title={Case 2: Datapath counter logic -- \texttt{Prob141\_count\_clock}}]
\textbf{Prompt.} A 12-hour BCD clock with reset priority, enable gating,
carry propagation, and AM/PM rollover.
\begin{lstlisting}[style=logexcerpt]
log: compile=pass, lint=pass, sim=fail, synth/pnr/gds=pass
     177947 mismatches in 200000 samples
tb:  Hint: reset value / reset priority
     Minute roll-over; Hour roll-over; PM roll-over
\end{lstlisting}
\end{tcolorbox}

\begin{tcolorbox}[casebox,title={Case 3: FSM corner case -- \texttt{Prob155\_lemmings4}}]
\textbf{Prompt.} A Lemmings FSM with walking, falling, digging, reset, and
long-fall splatter behavior.
\begin{lstlisting}[style=logexcerpt]
log: compile=pass, lint=pass, sim=fail, synth/pnr/gds=pass
     44 mismatches in 1003 samples
tb:  repeat(21) @(posedge clk);  // splat after falling left
     repeat(24) @(posedge clk);  // long-fall boundary
\end{lstlisting}
\end{tcolorbox}

\begin{tcolorbox}[casebox,title={Case 4: Interface artifact -- \texttt{Prob099\_m2014\_q6c}}]
\textbf{Prompt.} One-hot FSM next-state logic with benchmark-level port-name
inconsistency.
\begin{lstlisting}[style=logexcerpt]
log: compile=pass, lint=pass, sim=error
tb:  Prob099_m2014_q6c_test.sv:71: port `Y2' is not a port of good1.
     Prob099_m2014_q6c_test.sv:71: port `Y4' is not a port of good1.
     Prob099_m2014_q6c_test.sv:77: port `Y2' is not a port of top_module1.
\end{lstlisting}
\end{tcolorbox}

These cases show that the residual failures are not backend failures: the
designs either reach physical implementation and fail the behavioral oracle, or
fail at the benchmark harness interface before simulation can run.

\subsection{Example Physical Layouts}
\label{app:gds_example}

\begin{figure}[H]
\centering
\begin{subfigure}[t]{0.46\linewidth}
\centering
\includegraphics[width=\linewidth]{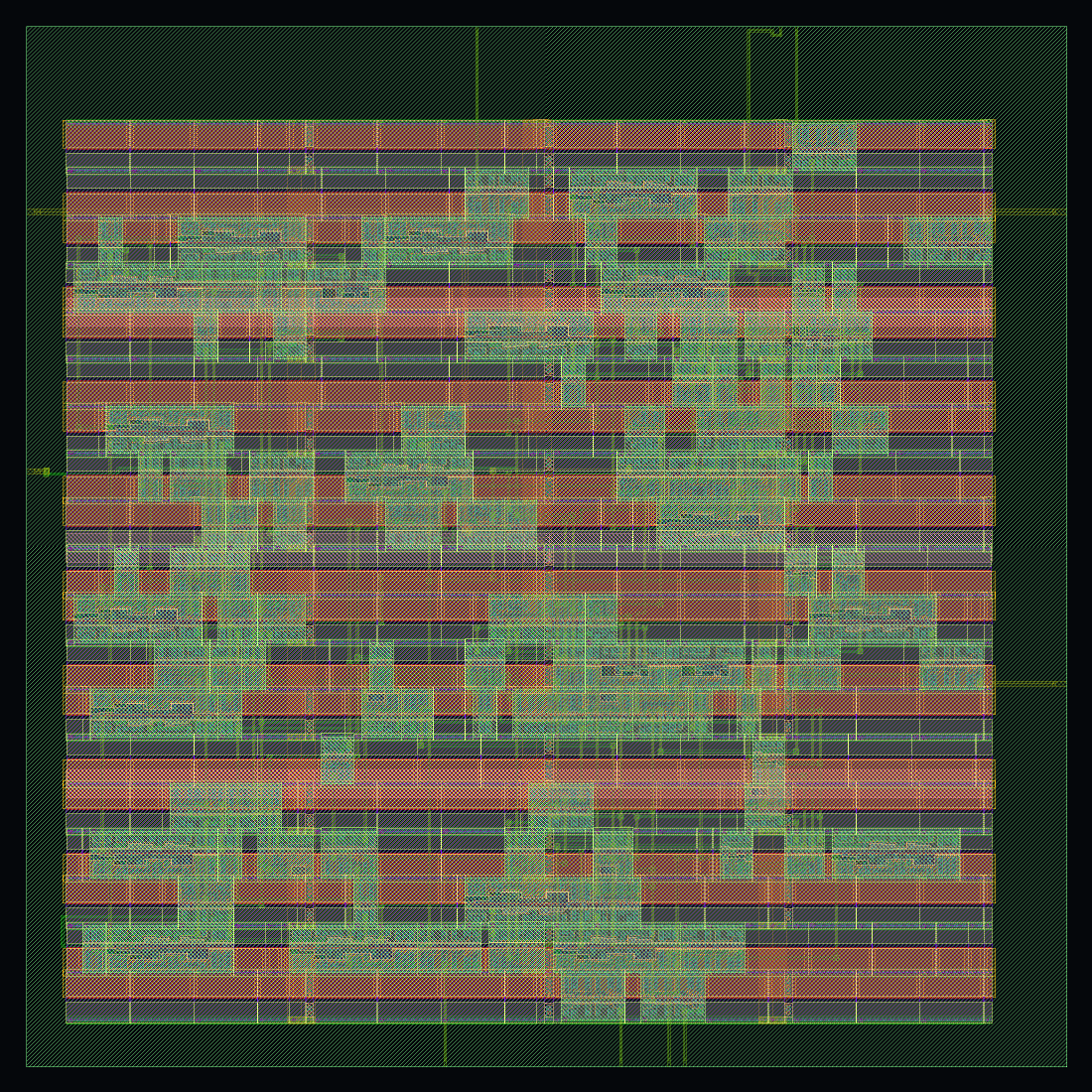}
\caption{\texttt{Prob156} (\emph{fancytimer})}
\label{fig:gds_prob156}
\end{subfigure}\hfill
\begin{subfigure}[t]{0.46\linewidth}
\centering
\includegraphics[width=\linewidth]{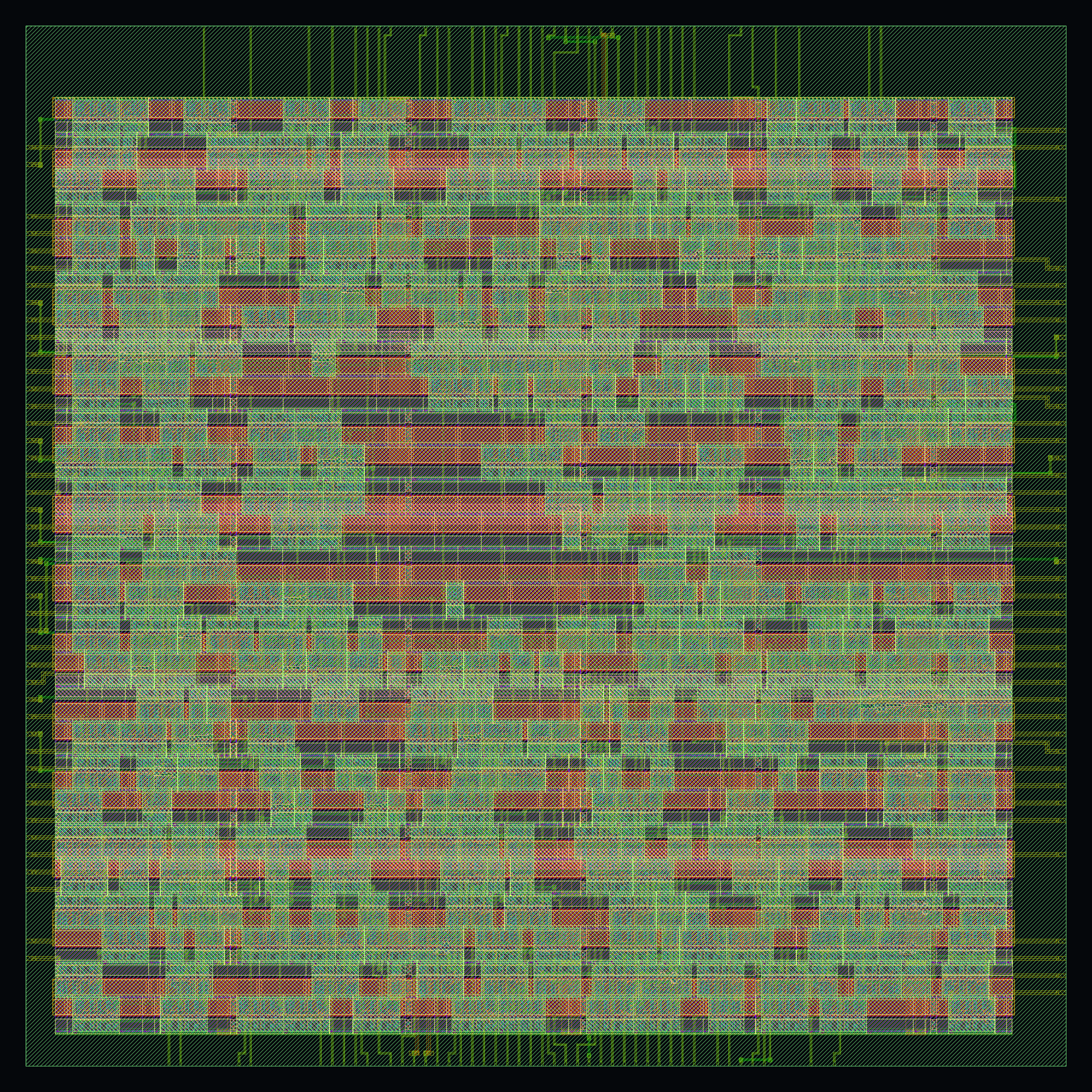}
\caption{\texttt{Prob097} (\emph{mux9to1v})}
\label{fig:gds_prob097}
\end{subfigure}
\caption{Raster renderings of OpenROAD-generated GDSII layouts for two VerilogEval designs.}
\label{fig:gds_examples}
\end{figure}

Figure~\ref{fig:gds_examples} shows two raster renderings of
OpenROAD-generated GDSII layouts for generated VerilogEval designs:
\texttt{Prob156} (\emph{fancytimer}), a multi-output sequential timer with BCD display logic,
and \texttt{Prob097} (\emph{mux9to1v}), a 9-to-1 multiplexer with 16-bit datapath.
These examples are included as qualitative illustrations of the layouts emitted by the automated synthesis and
place-and-route flow; quantitative backend pass rates are reported in
Section~\ref{sec:experiments}.

\subsection{AXI4-Lite Physical Layouts}
\label{app:axi4lite_layouts}

To demonstrate that \textsc{CktFormalizer} scales beyond educational benchmarks to industry-relevant IP, we generate both sides of an AXI4-Lite bus interface, the ARM AMBA protocol used for register access in virtually all modern SoCs. AXI4-Lite requires correct multi-channel handshake logic (AW/W/B for writes, AR/R for reads) with back-pressure and response sequencing; in hand-written Verilog, missing handshake cases and width mismatches across channels are a common source of protocol violations. In Lean HDL, the type system statically enforces channel widths and exhaustive case coverage, eliminating these error classes at compile time.

\begin{figure}[H]
\centering
\begin{subfigure}[t]{0.46\linewidth}
\centering
\includegraphics[width=\linewidth]{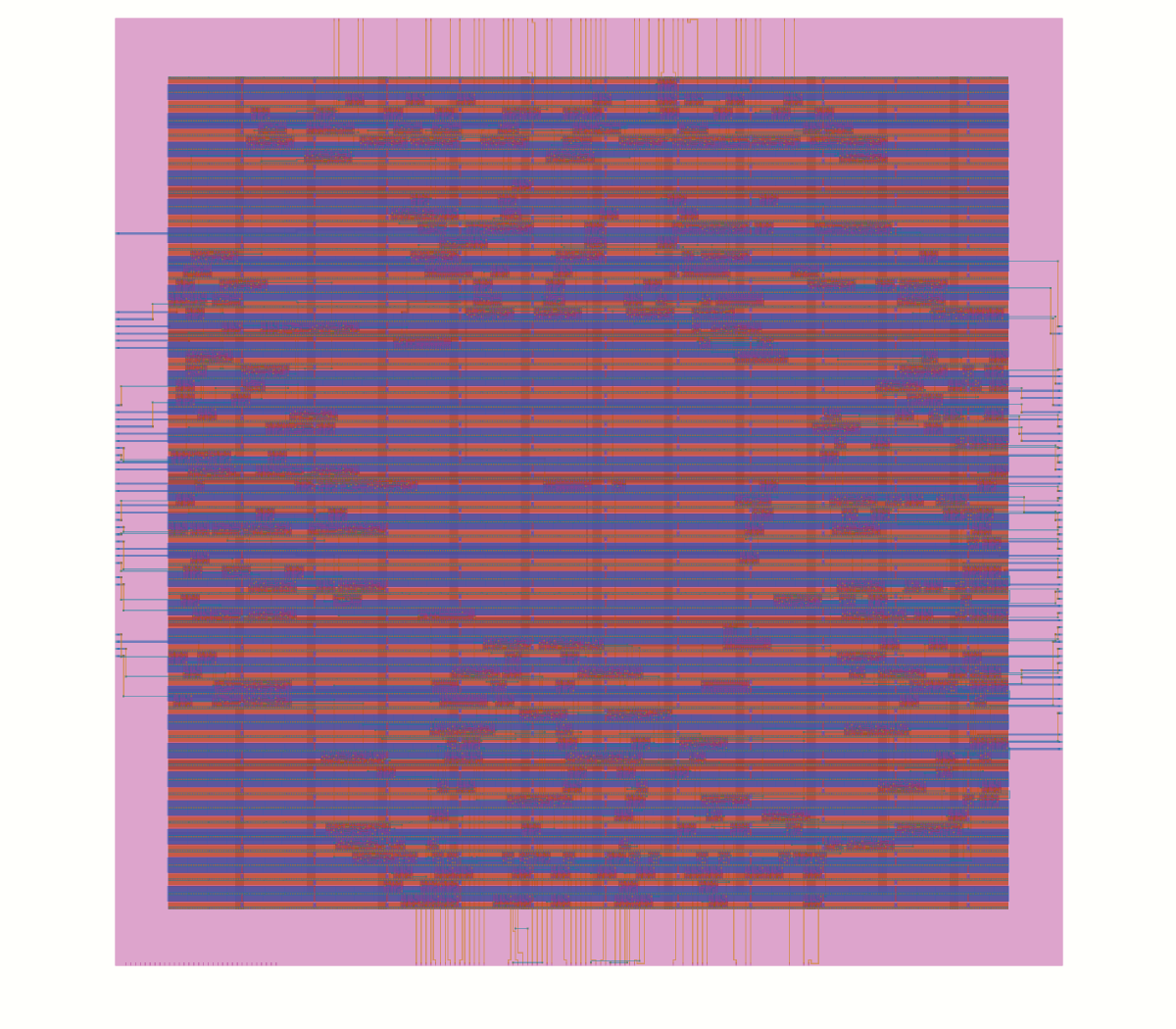}
\caption{Target-side module}
\label{fig:axi4lite_slave}
\end{subfigure}\hfill
\begin{subfigure}[t]{0.46\linewidth}
\centering
\includegraphics[width=\linewidth]{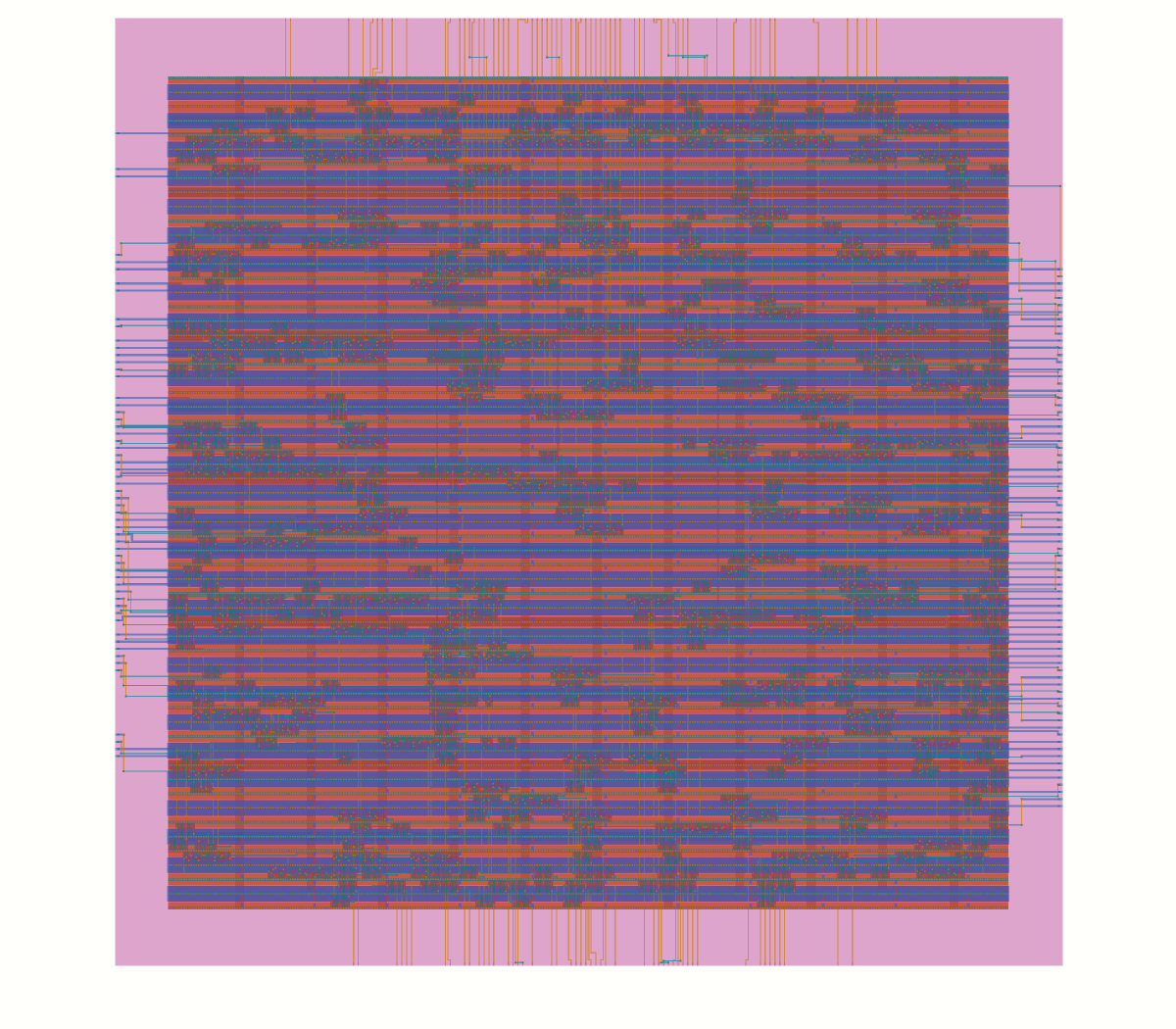}
\caption{Initiator-side module}
\label{fig:axi4lite_master}
\end{subfigure}
\caption{OpenROAD-generated AXI4-Lite target-side and initiator-side physical layouts.}
\label{fig:axi4lite_layouts}
\end{figure}

Figure~\ref{fig:axi4lite_layouts} shows the OpenROAD-generated layouts for the target-side (slave) and initiator-side (master) modules. Both designs pass the full automated pipeline, from natural language specification through Lean compilation, SystemVerilog extraction, synthesis, and place-and-route, with zero DRC, setup, hold, slew, fanout, and capacitance violations, requiring no manual intervention at any stage.

\clearpage
\section{Compact Result Dashboard}
\label{app:dashboard}

\begin{figure}[t]
\centering
\includegraphics[width=0.43\linewidth]{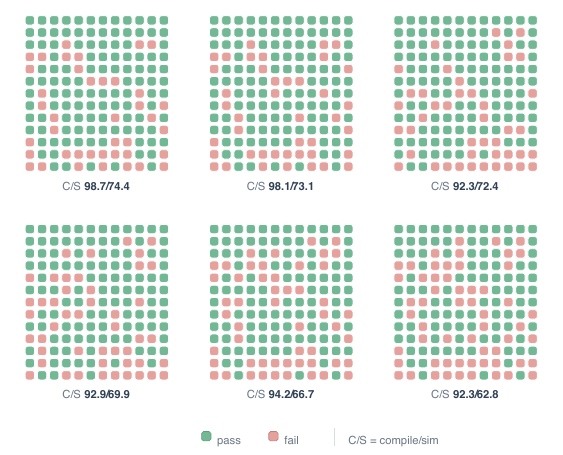}
\caption{Six compact pass/fail dashboards.}
\label{fig:dashboard}
\end{figure}

In addition to the aggregate metrics in Section~\ref{sec:experiments}, we use compact dashboards to inspect representative \textsc{CktFormalizer} runs at problem granularity. As shown in Figure~\ref{fig:dashboard}, the embedded view keeps only the audit signals needed for the paper: per-problem simulation outcomes and compact compile/simulation pass-rate summaries. The matrices make it easy to see whether failures are isolated to a few difficult problems or spread broadly across the benchmark.

\section{CktLean Implementation Details}
\label{app:cktlean_extensions}
\textsc{CktLean} extends \textsc{Sparkle}'s typed IR, elaboration, and backend lowering
to address the five capability gaps in \S\ref{sec:sparkle_limitations}.
The extensions below follow the same five categories and provide the
hardware infrastructure used by \textsc{CktFormalizer}.
Table~\ref{tab:capability_task_details} lists the 17 tasks and the capabilities they exercise.
\begin{table}[!htbp]
\centering
\caption{The 17 CVDP tasks used to analyze capability gaps in the original \textsc{Sparkle} toolchain. Task IDs omit the common \texttt{cvdp\_copilot\_} prefix. The last column identifies the capability exercised, rather than asserting an identical observed failure for every task.}
\label{tab:capability_task_details}
\small
\setlength{\tabcolsep}{4pt}
\renewcommand{\arraystretch}{1.12}
\begin{tabular}{@{}p{0.40\linewidth}p{0.18\linewidth}p{\dimexpr0.42\linewidth-16pt\relax}@{}}
\toprule
\textbf{Task ID} & \textbf{Category} & \textbf{Capability limitation examined} \\
\midrule
\texttt{car\_\allowbreak parking\_\allowbreak management\_\allowbreak 0001} & Stateful & Symbolic state widths and parameter dependencies. \\
\texttt{restoring\_\allowbreak division\_\allowbreak 0001} & Stateful & Derived widths, tuple state, and parameterized iteration. \\
\midrule
\texttt{sync\_\allowbreak lifo\_\allowbreak 0001} & Memory & Parameterized memory dimensions and read/write lowering. \\
\texttt{cache\_\allowbreak lru\_\allowbreak 0022} & Memory & Register-file reset and conditional write behavior. \\
\midrule
\texttt{hamming\_\allowbreak code\_\allowbreak tx\_\allowbreak and\_\allowbreak rx\_\allowbreak 0009} & Bit operations & Symbolic indexed bit construction and parity reduction. \\
\texttt{hamming\_\allowbreak code\_\allowbreak tx\_\allowbreak and\_\allowbreak rx\_\allowbreak 0011} & Bit operations & Symbolic indexed bit construction and parity reduction. \\
\midrule
\texttt{16qam\_\allowbreak mapper\_\allowbreak 0001} & Signed arithmetic & Signed/DSP representation and lowering. \\
\texttt{16qam\_\allowbreak mapper\_\allowbreak 0006} & Signed arithmetic & Signed/DSP representation and lowering. \\
\texttt{comparator\_\allowbreak 0001} & Signed arithmetic & Signed comparison semantics. \\
\texttt{complex\_\allowbreak multiplier\_\allowbreak 0001} & Signed arithmetic & Signed products and intermediate-width handling. \\
\texttt{configurable\_\allowbreak digital\_\allowbreak low\_\allowbreak pass\_\allowbreak filter\_\allowbreak 0001} & Signed arithmetic & Signed filtering arithmetic and width handling. \\
\texttt{configurable\_\allowbreak digital\_\allowbreak low\_\allowbreak pass\_\allowbreak filter\_\allowbreak 0014} & Signed arithmetic & Signed filtering arithmetic and width handling. \\
\texttt{sobel\_\allowbreak filter\_\allowbreak 0011} & Signed arithmetic & Signed intermediate values in filtering arithmetic. \\
\texttt{cont\_\allowbreak adder\_\allowbreak 0023} & Signed arithmetic & Signed addition and width handling. \\
\midrule
\texttt{cdc\_\allowbreak pulse\_\allowbreak synchronizer\_\allowbreak 0004} & Multiple clocks & Distinct clock domains and explicit pulse crossings. \\
\texttt{cdc\_\allowbreak pulse\_\allowbreak synchronizer\_\allowbreak 0013} & Multiple clocks & Distinct clock domains and explicit pulse crossings. \\
\texttt{fifo\_\allowbreak async\_\allowbreak 0001} & Multiple clocks & Independent read/write clocks and FIFO crossings. \\
\bottomrule
\end{tabular}
\end{table}

\vspace{-4mm}
\paragraph{Preserving parameterized state.}
We extend the typed IR with symbolic width and parameter expressions,
preserve derived parameters during elaboration, and propagate their
dependencies across module boundaries. Stateful circuits can thereby retain
parameter relationships through Lean elaboration, CppSim generation, and
SystemVerilog extraction.
\vspace{-4mm}
\paragraph{Lowering parameterized memories.}
We preserve symbolic memory dimensions and parameter-dependent read/write
structure in the IR and lowering pipeline. This allows memory declarations,
accesses, and generated RTL to share the same parameterization rather than
specializing memory structure prematurely. Behavioral checks remain
necessary for reset, initialization, and read/write interactions.
\vspace{-3mm}
\paragraph{Compiling generic bit operations.}
We add compiler support for symbolic per-bit structures and lower them
through parameter-aware generate constructs. Symbolic dimensions and local
aliases are preserved during elaboration so that indexed layouts are not
incorrectly collapsed into scalar or fixed-width forms.

\paragraph{Supporting signed arithmetic.}
We add signed signal and RTL helpers and extend the IR, elaborator, CppSim
backend, and SystemVerilog backend to preserve signedness, extension,
shifts, truncation, and parameterized widths consistently.

\paragraph{Representing multiple clock domains.}
We introduce first-class clock-domain information in the IR and associate
signals and state with their corresponding domains. The elaborator and
backends propagate this information through multi-domain circuits and
support lowering the corresponding CDC structures to CppSim and
SystemVerilog.

\subsection{Additional Typed Hardware Examples}
\label{app:typed_hdl_examples}
These examples complement Figure~\ref{fig:lean_hdl_examples}. Lean rejects
implicit self-reference as an unknown identifier; the supported generation
profile expresses feedback using explicit \texttt{Signal.reg} constructs.
The multiplexer covers every selector value, including a default branch;
omitting that branch leaves the pattern match incomplete and causes a compile error.

\begin{lstlisting}[style=lean4,caption={Implicit self-reference and exhaustive pattern matching in Lean HDL.},label={lst:typed_hdl_examples}]
-- Example 1: Implicit self-reference rejected by Lean
-- Feedback in the supported profile uses explicit Signal.reg constructs:
def bad_loop (x : Signal dom (BitVec 8)) : Signal dom (BitVec 8) :=
  let y := x + y  -- ERROR: unknown identifier 'y'
  y

-- Example 2: Exhaustive pattern matching prevents unintended latches
def mux3 (sel : Signal dom (BitVec 2))
         (a b c : Signal dom (BitVec 8)) : Signal dom (BitVec 8) :=
  Signal.match sel fun
    | 0b00 => a
    | 0b01 => b
    | 0b10 => c
    | _    => 0#8  -- must cover all cases; omitting causes compile error
\end{lstlisting}

\section{Formal Verification and Equivalence Checking}
\label{app:formal_verification}

Lean HDL supports hardware descriptions and formal proofs within the same
language. Signals have a denotational semantics as functions over time
($\texttt{Signal dom }\alpha\triangleq\mathbb{N}\to\alpha$), allowing behavioral
properties and implementation equivalences to be expressed as Lean theorems.
Completed proofs are checked by the Lean kernel. We complement these
source-level proofs with RTL equivalence checking and simulation, each
providing evidence at a different abstraction level.

\paragraph{Property and Equivalence Proofs.}
Designers or the LLM agent can state behavioral properties directly over signal
definitions, such as a counter's initial value and increment invariant
(Figure~\ref{fig:formal_verification}, top). Given a specification \texttt{spec}
and an implementation \texttt{opt}, the agent can also attempt to prove
$\forall\,\mathit{input}.\;\texttt{opt}(\mathit{input})=\texttt{spec}(\mathit{input})$
as a Lean theorem (Figure~\ref{fig:formal_verification}, bottom).

When formal verification is enabled during architecture exploration or PPA
optimization (\S\ref{sec:agent_workflow}), the agent uses a ReAct-style loop
to formalize selected properties from the natural-language specification and
iteratively prove that the implementation satisfies them. The persistent Lean
REPL serves as an interactive interface for stepwise proof construction: the
agent submits a proof step, inspects the resulting goals and diagnostics, and
uses this feedback to select or revise the next step. This interaction connects
the agent's reasoning to the evolving Lean proof state throughout the proof
attempt. A completed, kernel-accepted proof establishes the stated source-level
property or equivalence, but does not by itself establish that the formal
specification faithfully captures the natural-language request. An unfinished
proof provides no such certificate, but does not by itself demonstrate a
functional defect.

For the signal-equality theorem shown in Figure~\ref{fig:formal_verification}, a
completed proof establishes equality for all input signals and all time steps
within the modeled synchronous semantics. This guarantee extends beyond the
finite traces exercised by simulation, while remaining specific to the formal
specification and signal model.

\begin{figure}[!t]
\centering
\begin{lstlisting}[style=lean4]
-- Property proofs: counter initial value and increment invariant
theorem counter_starts_at_zero :
    (counter.atTime 0) = 0#8 := by rfl

theorem counter_increments (n : Nat) :
    (counter.atTime (n+1)) = (counter.atTime n) + 1#8 := by
  simp [counter, Signal.register]; rfl

-- Equivalence proof: optimized implementation = specification
theorem equiv (input : Signal dom (BitVec 3)) :
    optimized input = spec input := by
  unfold optimized spec; rfl
\end{lstlisting}
\caption{Formal verification in Lean HDL. Using the counter defined in Figure~\ref{fig:lean_hdl_examples}, the top theorems prove behavioral properties (initial value and increment invariant). The bottom theorem establishes equality between an optimized implementation and its specification for all input signals and time steps within the modeled synchronous semantics.}
\label{fig:formal_verification}
\end{figure}
\paragraph{RTL Equivalence Checking.}
To complement source-level proofs, we use Yosys equivalence checking to compare
the generated SystemVerilog against the VerilogEval reference implementation.
The checking procedure combines SAT-based combinational reasoning
(\texttt{equiv\_simple}) with sequential induction (\texttt{equiv\_induct}).
Successful checks establish RTL equivalence under the procedure's modeling
assumptions, independently of whether a Lean proof is available.

Proof automation remains design-dependent at both levels. Lean tactics such as
\texttt{simp} and \texttt{decide} can discharge suitable obligations, while more
complex sequential designs may require additional lemmas or induction arguments
that the agent does not always construct successfully. Similarly, Yosys checks
may remain inconclusive for designs beyond the capabilities of the configured
reduction and induction procedures, including deeply pipelined or memory-heavy
designs. These limitations concern the ability to complete verification, rather
than weakening the guarantee of a successfully completed proof.

\begin{table}[!htbp]
\centering
\small
\caption{Evidence provided by the checking mechanisms used in the workflow.
Proof or equivalence guarantees apply only when the corresponding check
completes successfully.}
\label{tab:equiv_scope}
\setlength{\tabcolsep}{4pt}
\begin{tabular}{@{}p{0.19\linewidth}p{0.36\linewidth}p{\dimexpr0.45\linewidth-16pt\relax}@{}}
\toprule
\textbf{Mechanism} & \textbf{Evidence established} & \textbf{Scope and limitations} \\
\midrule
Lean type checking & Supported structural invariants, including bit-width consistency and exhaustive case coverage & Does not by itself establish functional correctness \\
\addlinespace
Lean kernel-checked proofs & Stated behavioral properties or source-level equivalence & Applies within the formal signal model; proof automation is design-dependent \\
\addlinespace
Yosys equivalence checking & Equivalence between emitted RTL and the benchmark reference under the checking assumptions & Limited by supported constructs and the configured proof procedures \\
\addlinespace
RTL simulation & Agreement with expected behavior on exercised traces, including supported reset and memory scenarios & Finite test coverage does not establish universal equivalence \\
\bottomrule
\end{tabular}
\end{table}

\paragraph{Guarantees and Verification Boundaries.}
Table~\ref{tab:equiv_scope} separates the evidence provided by each checking
mechanism from its scope and limitations. Lean type checking establishes
supported structural invariants, such as bit-width consistency and exhaustive
case coverage. Completed Lean theorems establish the source-level functional
properties they state. Yosys checks compare the emitted RTL with the benchmark
reference, while simulation assesses behavior on finite testbench traces.

A central verification boundary lies between the Lean source semantics and the
emitted SystemVerilog. Lean proofs concern source-level definitions, whereas
Yosys checks concern the emitted RTL and a separate reference implementation.
Neither directly establishes that the Lean-to-SystemVerilog translation
preserves the source semantics. The current pipeline therefore does not provide
an end-to-end proof of compiler correctness.

Two further categories remain outside the current formal verification scope.
First, the signal model represents behavior within a single synchronous clock
domain. Multi-clock interactions and asynchronous reset behavior lack a formal
semantics in the current framework; verifying these behaviors requires an
extended semantic model or complementary analyses, such as simulation and
static clock-domain-crossing checks. Second, the current checks do not establish
timing-aware or gate-level equivalence. Post-synthesis transformations, cell
timing, and physical parasitics require additional verification beyond the
source-level proofs and RTL checks used here.

\section{Architecture Exploration Data}
\label{app:arch_explore_data}
Table~\ref{tab:arch_designs} expands the abbreviated design labels used in
Table~\ref{tab:arch_explore_reductions}. These descriptions explain the
labels; establishing the exact benchmark task and variant for each row
requires the original per-problem architecture-exploration records.
\begin{table}[htbp]
\centering
\small
\caption{Meaning of the design labels used in Table~\ref{tab:arch_explore_reductions}. Descriptions identify circuit categories; exact benchmark task IDs are not available in the archived summary.}
\label{tab:arch_designs}
\setlength{\tabcolsep}{5pt}
\renewcommand{\arraystretch}{1.15}
\begin{tabular}{@{}p{0.24\linewidth}p{0.18\linewidth}p{\dimexpr0.58\linewidth-20pt\relax}@{}}
\toprule
\textbf{Table label} & \textbf{Circuit type} & \textbf{Functional interpretation} \\
\midrule
FSM (Lemmings) & Sequential control & Finite-state controller for a Lemmings task. The summary does not identify which Lemmings variant was evaluated. \\
9-to-1 Mux & Combinational & Multiplexer selecting one of nine data inputs for its output. \\
Add/Sub Unit & Arithmetic & Datapath supporting addition and subtraction. \\
FSM Shift Reg. & Sequential control & Finite-state control associated with shift-register operation. \\
Case-z Logic & Combinational & Selection logic expressed using wildcard case matching (\texttt{casez}). \\
Comb.\ Logic & Combinational & Logic whose outputs depend on current inputs; the label does not specify the particular Boolean function. \\
\bottomrule
\end{tabular}
\end{table}

Table~\ref{tab:arch_explore} provides the absolute synthesis areas, design types,
and reduction percentages for the six examples in Table~\ref{tab:arch_explore_reductions}.
The examples are drawn from simulation-passing candidates and do not represent
all problems in the architecture exploration experiment.
\begin{table}[!htbp]
\centering
\small
\caption{Architecture exploration data for the six simulation-passing examples in Table~\ref{tab:arch_explore_reductions}. Initial and candidate areas are in $\mu$m$^2$; the final three columns report reductions.}
\label{tab:arch_explore}
\setlength{\tabcolsep}{2pt}
\begin{tabular*}{\linewidth}{@{\extracolsep{\fill}}llrrrrr@{}}
\toprule
\textbf{Design} & \textbf{Type} & \textbf{Init. area} & \textbf{Cand. area} & \textbf{Area} & \textbf{Cell} & \textbf{Power} \\
\midrule
FSM (Lemmings) & Seq. & 198.9 & 128.9 & \textbf{35.2\%} & 35.0\% & 30.3\% \\
9-to-1 Mux & Comb. & 1{,}818.0 & 1{,}321.3 & \textbf{27.3\%} & 45.9\% & 14.5\% \\
Add/Sub Unit & Arith. & 483.0 & 399.1 & \textbf{17.4\%} & 26.2\% & 8.9\% \\
FSM Shift Reg. & Seq. & 131.4 & 116.4 & \textbf{11.4\%} & 0.0\% & 7.5\% \\
Case-z Logic & Comb. & 83.8 & 75.1 & \textbf{10.4\%} & 0.0\% & 2.0\% \\
Comb.\ Logic & Comb. & 226.5 & 203.9 & \textbf{9.9\%} & 14.8\% & 6.4\% \\
\bottomrule
\end{tabular*}
\end{table}

\section{Mounted In-Context Idiom Files}
\label{app:idiom_library}

This section lists the 13 Lean files mounted for \textsc{CktFormalizer}
as style references (Appendix~\ref{app:benchmarks}). Each file begins with
\texttt{import Sparkle}, \texttt{import Sparkle.Compiler.Elab}, and
\texttt{open Sparkle.Core.\{Domain,Signal\}}. Gold Verilog from benchmark
comments is omitted.
\newpage
\begin{lstlisting}[style=lean4,caption={\texttt{Prob005\_notgate.lean}},label={lst:idiom_notgate}]
def prob005_notgate {dom : DomainConfig}
    (input : Signal dom (BitVec 1)) : Signal dom (BitVec 1) :=
  ~~~input
#synthesizeVerilog prob005_notgate
\end{lstlisting}
\begin{lstlisting}[style=lean4,caption={\texttt{Prob022\_mux2to1.lean}},label={lst:idiom_mux}]
def prob022_mux2to1 {dom : DomainConfig}
    (a b : Signal dom (BitVec 1)) (sel : Signal dom Bool)
    : Signal dom (BitVec 1) :=
  Signal.mux sel b a
#synthesizeVerilog prob022_mux2to1
\end{lstlisting}
\begin{lstlisting}[style=lean4,caption={\texttt{Prob024\_hadd.lean}},label={lst:idiom_hadd}]
def prob024_hadd {dom : DomainConfig}
    (a b : Signal dom (BitVec 1))
    : Signal dom (BitVec 1 × BitVec 1) :=
  let sum  := a ^^^ b
  let cout := a &&& b
  bundle2 sum cout
#synthesizeVerilog prob024_hadd
\end{lstlisting}
\begin{lstlisting}[style=lean4,caption={\texttt{Prob027\_fadd.lean}},label={lst:idiom_fadd}]
def prob027_fadd {dom : DomainConfig}
    (a b cin : Signal dom (BitVec 1))
    : Signal dom (BitVec 1 × BitVec 1) :=
  let sum  := a ^^^ b ^^^ cin
  let cout := (a &&& b) ||| (a &&& cin) ||| (b &&& cin)
  bundle2 sum cout
#synthesizeVerilog prob027_fadd
\end{lstlisting}
\begin{lstlisting}[style=lean4,caption={\texttt{Prob031\_dff.lean}},label={lst:idiom_dff}]
def prob031_dff {dom : DomainConfig}
    (d : Signal dom (BitVec 1)) : Signal dom (BitVec 1) :=
  Signal.register 0#1 d
#synthesizeVerilog prob031_dff
\end{lstlisting}
\begin{lstlisting}[style=lean4,caption={\texttt{Prob035\_count1to10.lean}},label={lst:idiom_count}]
def prob035_count1to10 {dom : DomainConfig}
    (reset : Signal dom Bool) : Signal dom (BitVec 4) :=
  Signal.loop fun (q : Signal dom (BitVec 4)) =>
    let atMax := q === 10#4
    let shouldReset := reset ||| atMax
    let nextVal := Signal.mux shouldReset (Signal.pure 1#4) (q + 1#4)
    Signal.register 1#4 nextVal
#synthesizeVerilog prob035_count1to10
\end{lstlisting}
\begin{lstlisting}[style=lean4,caption={\texttt{Prob054\_edgedetect.lean}},label={lst:idiom_edge}]
def prob054_edgedetect {dom : DomainConfig}
    (input : Signal dom (BitVec 8)) : Signal dom (BitVec 8) :=
  let d_last := Signal.register 0#8 input
  let edges := input &&& (~~~d_last)
  Signal.register 0#8 edges
#synthesizeVerilog prob054_edgedetect
\end{lstlisting}
\begin{lstlisting}[style=lean4,caption={\texttt{Prob086\_lfsr5.lean}},label={lst:idiom_lfsr}]
def prob086_lfsr5 {dom : DomainConfig}
    (reset : Signal dom Bool) : Signal dom (BitVec 5) :=
  Signal.loop fun (q : Signal dom (BitVec 5)) =>
    let shifted := q >>> 1#5
    let lsb := q &&& (1#5 : BitVec 5)
    let fb := lsb === (1#5 : BitVec 5)
    let q_next := Signal.mux fb (shifted ^^^ (20#5 : BitVec 5)) shifted
    let nextVal := Signal.mux reset (Signal.pure 1#5) q_next
    Signal.register 1#5 nextVal
#synthesizeVerilog prob086_lfsr5
\end{lstlisting}
\begin{lstlisting}[style=lean4,caption={\texttt{Prob091\_fsm\_logic.lean}},label={lst:idiom_fsm_logic}]
def prob091_fsm_logic {dom : DomainConfig}
    (y : Signal dom (BitVec 6)) (w : Signal dom Bool)
    : Signal dom (BitVec 1 × BitVec 1) :=
  let y0 := Signal.map (fun x => x.getLsb 0) y
  let y1 := Signal.map (fun x => x.getLsb 1) y
  let y2 := Signal.map (fun x => x.getLsb 2) y
  let y4 := Signal.map (fun x => x.getLsb 4) y
  let y5 := Signal.map (fun x => x.getLsb 5) y
  let Y1_bool := y0 &&& w
  let Y1 := Signal.mux Y1_bool (Signal.pure 1#1) (Signal.pure 0#1)
  let states_to_D := y1 ||| y2 ||| y4 ||| y5
  let Y3_bool := states_to_D &&& (~~~w)
  let Y3 := Signal.mux Y3_bool (Signal.pure 1#1) (Signal.pure 0#1)
  bundle2 Y1 Y3
#synthesizeVerilog prob091_fsm_logic
\end{lstlisting}
\begin{lstlisting}[style=lean4,caption={\texttt{Prob109\_fsm1.lean}},label={lst:idiom_fsm1}]
private abbrev stA : BitVec 1 := 0#1
private abbrev stB : BitVec 1 := 1#1
def prob109_fsm1 {dom : DomainConfig}
    (areset : Signal dom Bool) (inp : Signal dom Bool)
    : Signal dom (BitVec 1) :=
  Signal.loop fun (state : Signal dom (BitVec 1)) =>
    let stateFlipped := ~~~state
    let nextState := Signal.mux inp state stateFlipped
    let nextWithReset := Signal.mux areset (Signal.pure stB) nextState
    Signal.register stB nextWithReset
#synthesizeVerilog prob109_fsm1
\end{lstlisting}
\begin{lstlisting}[style=lean4,caption={\texttt{Prob127\_lemmings1.lean}},label={lst:idiom_lemmings}]
private abbrev stWL : BitVec 1 := 0#1
private abbrev stWR : BitVec 1 := 1#1
def prob127_lemmings1 {dom : DomainConfig}
    (areset : Signal dom Bool)
    (bump_left bump_right : Signal dom Bool)
    : Signal dom (BitVec 1 × BitVec 1) :=
  let state : Signal dom (BitVec 1) :=
    Signal.loop fun (state : Signal dom (BitVec 1)) =>
      let isWL := state === (Signal.pure stWL)
      let nextFromWL := Signal.mux bump_left (Signal.pure stWR) (Signal.pure stWL)
      let nextFromWR := Signal.mux bump_right (Signal.pure stWL) (Signal.pure stWR)
      let nextState := Signal.mux isWL nextFromWL nextFromWR
      let nextWithReset := Signal.mux areset (Signal.pure stWL) nextState
      Signal.register stWL nextWithReset
  bundle2 (~~~state) state
#synthesizeVerilog prob127_lemmings1
\end{lstlisting}
\begin{lstlisting}[style=lean4,caption={\texttt{RTLIdioms.lean}},label={lst:idiom_rtlidioms}]
def idiomSlicing {dom : DomainConfig}
    (x : Signal dom (BitVec 16))
    : Signal dom (BitVec 8 × BitVec 8 × BitVec 1) :=
  let lo : Signal dom (BitVec 8) := Sparkle.Library.RTL.slice x 0
  let hi : Signal dom (BitVec 8) := Sparkle.Library.RTL.slice x 8
  let nonzero := Sparkle.Library.RTL.boolToBV1 (Sparkle.Library.RTL.nonZero x)
  bundleAll! [hi, lo, nonzero]
#synthesizeVerilog idiomSlicing

def idiomReverse8 {dom : DomainConfig}
    (x : Signal dom (BitVec 8)) : Signal dom (BitVec 8) :=
  Sparkle.Library.RTL.reverseBits8 x
#synthesizeVerilog idiomReverse8

def idiomBitScans {dom : DomainConfig}
    (x : Signal dom (BitVec 8)) : Signal dom (BitVec 4 × BitVec 3) :=
  let count := Sparkle.Library.RTL.popCount8 x
  let firstSet := Sparkle.Library.RTL.priorityEncodeLsb8 x
  bundle2 count firstSet
#synthesizeVerilog idiomBitScans

def idiomRegFile {dom : DomainConfig}
    (wen : Signal dom Bool)
    (waddr : Signal dom (BitVec 4))
    (wdata : Signal dom (BitVec 8))
    (raddr : Signal dom (BitVec 4))
    : Signal dom (BitVec 8) :=
  Sparkle.Library.RTL.regFile1R1W waddr wdata wen raddr
#synthesizeVerilog idiomRegFile

def idiomCounterEnableReset {dom : DomainConfig}
    (rstN en : Signal dom Bool) : Signal dom (BitVec 8) :=
  Signal.loop fun q =>
    let inc := q + 1#8
    let next := Sparkle.Library.RTL.resetLow 0#8 rstN inc
    Sparkle.Library.RTL.dffe 0#8 en next
#synthesizeVerilog idiomCounterEnableReset
\end{lstlisting}
\begin{lstlisting}[style=lean4,caption={\texttt{TestClkParam.lean}},label={lst:idiom_clkparam}]
def test_clk_param {dom : DomainConfig}
    (clk : Signal dom Bool)
    (d   : Signal dom (BitVec 8))
    : Signal dom (BitVec 8) :=
  Signal.mux clk d (Signal.pure 0#8)
#synthesizeVerilog test_clk_param
\end{lstlisting}

\end{document}